\documentclass[10pt,twocolumn,letterpaper]{article}

\usepackage{iccv}
\usepackage{times}
\usepackage{epsfig}
\usepackage{graphicx}
\usepackage{amsmath}
\usepackage{amssymb}
\usepackage{capt-of}
\usepackage{caption}
\usepackage{booktabs}
\usepackage{dashrule}
\usepackage{cuted}
\usepackage[square,comma,sort&compress,numbers]{natbib}

\usepackage[pagebackref=true,breaklinks=true,letterpaper=true,colorlinks,bookmarks=false]{hyperref}

\iccvfinalcopy 

\def\iccvPaperID{4637} 

\ificcvfinal\fi

\begin{document}

\title{3D-aware Image Generation using 2D Diffusion Models}

\author{Jianfeng Xiang$^{1,2}$\ \!\thanks{Work done when JX and BH were interns at MSR.} \quad Jiaolong Yang$^{2}$ \quad Binbin Huang$^{3}$ \quad Xin Tong$^{2}$ \\
	$^1${Tsinghua University} \quad  $^2${Microsoft Research Asia} \quad $^3${ShanghaiTech University} \\
\tt{\url{https://jeffreyxiang.github.io/ivid/}}
}

\maketitle
\ificcvfinal\thispagestyle{empty}\fi

\begin{strip}
	\vspace{-48pt}
	\centering
	\includegraphics[width=1\linewidth]{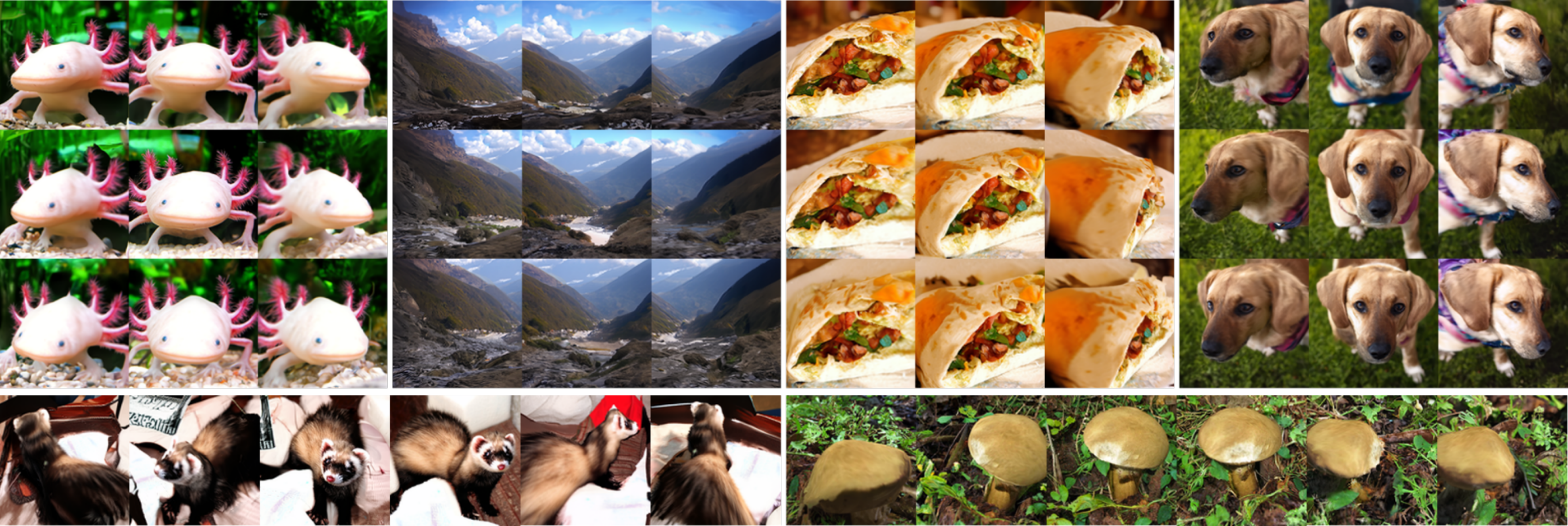}
	\vspace{-19pt}
	\captionsetup{type=figure,font=small}
	\caption{Our diffusion-based 3D-aware image generation trained on ImageNet. The first three rows shows the diverse objects and scenes generated by our method. The bottom row shows two cases synthesized under a 360$^\circ$ camera trajectory.}
 \label{fig:teaser}
	\vspace{0pt}
\end{strip}

\begin{abstract}
In this paper, we introduce a novel 3D-aware image generation method that leverages 2D diffusion models.
We formulate the 3D-aware image generation task as multiview 2D image set generation, and further to a sequential unconditional--conditional multiview image generation process. 
This allows us to utilize 2D diffusion models to boost the generative modeling power of the method. 
Additionally, we incorporate depth information from monocular depth estimators to construct the training data for the conditional diffusion model using only still images. 

We train our method on a large-scale dataset, i.e., ImageNet, which is not addressed by
previous methods. It produces high-quality images that significantly outperform prior methods. Furthermore, our approach showcases its capability to generate instances with large view angles, even though the training images are diverse and unaligned, gathered from ``in-the-wild" real-world environments. 
\vspace{-4pt}
\end{abstract}

\section{Introduction}
Learning to generate 3D assets has become an increasingly prominent task due to its numerous applications such as VR/AR, movie production, and art design.
Significant progress has been made recently in 3D-aware image generation, with a variety of approaches being proposed~\cite{schwarz2020graf, chan2021pi, deng2021gram, niemeyer2021giraffe,xiang2022gram, chan2022efficient, or2022stylesdf, gu2021stylenerf}. The goal of 3D-aware image generation is to train image generation models that are capable of explicitly controlling 3D camera pose, typically by using only unstructured 2D image collections.

Most existing methods for 3D-aware image generation rely on Generative Adversarial Networks (GANs)~\cite{goodfellow2014generative} and utilize a Neural Radiance Field (NeRF)~\cite{mildenhall2020nerf} or its variants as the 3D scene representation. While promising results have been demonstrated for object-level generation, extending these methods to large-scale, in-the-wild data that features significantly more complex variations in geometry and appearance remains a challenge. 

Diffusion Models (DMs)~\cite{sohl2015deep, ho2020denoising, song2020score}, on the other hand, are increasingly gaining recognition for their exceptional generative modeling performance on billion-scale image datasets~\cite{rombach2022high, ramesh2022hierarchical, saharia2022photorealistic}. It has been shown that DMs have surpassed GANs as the state-of-the-art models for complex image generation tasks~\cite{nichol2021improved, ho2022cascaded, dhariwal2021diffusion, ho2022classifier}. However, applying DMs to 3D-aware image generation tasks is not straightforward. One prominent hurdle is training data, as training DMs for 3D generation necessitates raw 3D assets for its nature of regression-based learning~\cite{mueller2022diffrf, luo2021diffusion, zhou20213d, nichol2022point, shue20223d}.

To take advantage of the potent capability of DMs and the ample  availability of 2D data, our core idea in this paper is to \emph{formulate 3D-aware generation as a multiview 2D image set generation task}.
Two critical issues must be addressed for this newly formulated task. The first is how to apply DMs for image \emph{set} generation. Our solution to this is to cast set generation as a \emph{sequential unconditional--conditional generation} process by factorizing the joint distribution of multiple views of an instance using the chain rule of probability. More specifically, we sample the initial view of an instance using an unconditional DM, followed by iteratively sampling other views with previous views as conditions via a conditional DM. 
This not only minimizes the model's output to a single image per generation, but also grants it the ability to handle variable numbers of output views.

The second issue is the lack of multiview image data. Inspired by a few recent studies~\cite{han2022single,cai2022diffdreamer}, we append \emph{depth} information to the image data through monocular depth estimation techniques and use depth to construct multiview data using only still images. However, we found that naively applying the data construction strategy of \cite{han2022single} can result in domain gaps between training and inference. To alleviate this, we recommend  additional training data augmentation strategies that can improve the generation quality, particularly for the results under large view angles.

We tested our method on both a large-scale, multi-class dataset, \ie, ImageNet~\cite{imagenet}, and several smaller, single-category datasets that feature significant variations in geometry. The results show that our method outperformed state-of-the-art 3D-aware GANs on ImageNet by a wide margin, demonstrating the significantly enhanced generative modeling capability of our novel 3D-aware generation approach. It also performed favorably against prior art on other datasets, showing comparable texture quality but improved geometry.
Moreover, we find that our model has the capability to generate scenes under large view angles (up to 360 degrees) from unaligned training data, which is a challenging task further demonstrating the efficacy of our new method.

\vspace{2pt}
\textbf{The contributions of this work} are summarized below:
\begin{itemize}
	\vspace{-3pt}
	\item We present a novel 3D-aware image generation method that uses 2D diffusion models. The method is designed based on a new formulation for 3D-aware generation, \ie, sequential unconditional--conditional multiview image sampling.
	\vspace{-3pt}
	\item We undertake 3D-aware generation on a large-scale in-the-wild dataset (ImageNet), which is not addressed by previous 3D-aware generation models. 
	\vspace{-3pt}
	\item We demonstrate the capability of our method for large-angle generation from unaligned data (up to 360 degrees).
\end{itemize}

\section{Related Work}

\paragraph{3D-aware image generation}

Previous 3D-aware image generation studies \cite{schwarz2020graf, chan2021pi, deng2021gram, niemeyer2021giraffe, chan2022efficient, or2022stylesdf, gu2021stylenerf} have achieved this objective on some well-aligned image datasets of specific objects.
Most of these works are based on GANs~\cite{goodfellow2014generative}. Some of them \cite{schwarz2020graf, chan2021pi, deng2021gram, xiang2022gram, skorokhodov2022epigraf} generate 3D scene representations which are used to directly render the final output images. They typically leverage NeRF~\cite{mildenhall2020nerf} or its variants as the 3D scene representation and train a scene generator with supervision on the rendered images from a jointly-trained discriminator. Others have combined 3D representation with 2D refinements \cite{niemeyer2021giraffe, gu2021stylenerf, or2022stylesdf, chan2022efficient}, performing two steps: generating a low-resolution volume to render 2D images or feature maps, and then refining the 2D images with a super-resolution module.
Very recently, two works concurrent to us \cite{skorokhodov3d, sargent2023vq3d} expand 3D-aware generation task to large and diverse 2D image collections such as ImageNet~\cite{imagenet}, utilizing geometric priors from pretrained monocular depth prediction models. This work presents a novel 2D diffusion based 3D-aware generative model, which can be applied to diverse in-the-wild 2D images.

\vspace{-10pt}
\paragraph{Diffusion models}

Diffusion models~\cite{sohl2015deep} come with a well-conceived theoretical formulation and U-net architecture, making them suitable for image modeling tasks~\cite{ho2020denoising, song2020score}. 
Improved diffusion-based methods \cite{nichol2021improved, ho2022cascaded, dhariwal2021diffusion, ho2022classifier} demonstrated that DMs have surpassed GANs as the new state-of-the-art models for some image generation tasks. Additionally, diffusion models can be applied to conditional generation, leading to the flourishing of downstream image-domain tasks such as image super-resolution~\cite{saharia2022image, li2022srdiff}, inpainting~\cite{lugmayr2022repaint, saharia2022palette, rombach2022high}, novel view synthesis~\cite{watson2022novel} and scene synthesis~\cite{cai2022diffdreamer,lei2022generative}. Our method utilizes 2D unconditional and conditional diffusion models with an iterative view sampling process to tackle 3D-aware generation.

\vspace{-10pt}
\paragraph{Optimization-based 3D generation}
According to the theory of diffusion models, the U-nets are trained to the \emph{score function} (log derivative) of the image distribution under different noise levels~\cite{song2020score}. 
This has led to the development of the Score Distillation Sampling (SDS) technique, which has been used to perform text-to-3D conversion using a text-conditioned diffusion model, with SDS serving as the multiview objective to optimize a NeRF-based 3D representation.
Although recent works \cite{wang2022score, lin2022magic3d} have explored this technique on different diffusion models and 3D representations, they are not generative models and are not suitable for random generation without text prompt.

\vspace{-10pt}
\paragraph{Depth-assisted view synthesis} Some previous works utilized depth information for view synthesis tasks including single-view view synthesis~\cite{han2022single, niklaus20193d} and perpetual view generation~\cite{liu2021infinite, li2022infinitenature, cai2022diffdreamer}. 
In contrast, this work deals with a different task, \ie, 3D-aware generative modeling of 2D image distributions. For our task, we propose a new formulation of sequential unconditional--conditional multiview image sampling, where the latter conditional generation subroutine shares a similar task with novel view synthesis.

\section{Problem Formulation}
\subsection{Preliminaries}

In this section, we provide a brief overview of the theory behind Diffusion Models and Conditional Diffusion Models~\cite{ho2020denoising, song2020score}. DMs are probabilistic generative models that are designed to recover images from a specified degradation process. To achieve this, two Markov chains are defined. The \emph{forward chain} is a destruction process that progressively adds Gaussian noise to target images:
\begin{equation}
    q(\bold x_t|\bold x_{t-1})=\mathcal{N}(\bold x_t;\sqrt{1-\beta_t}\bold x_{t-1},\beta_t\bold I).
\end{equation}
This process results in the complete degradation of target images in the end, leaving behind only tractable Gaussian noise. The \emph{reverse chain} is then employed to iteratively recover images from noise:
\begin{equation}
    p_\theta(\bold x_{t-1}|\bold x_t)=\mathcal{N}(\bold x_{t-1};\mu_\theta(\bold x_t,t);\Sigma_\theta(\bold x_t,t)),
\end{equation}
where the mean and variance functions are modeled as neural networks trained by minimizing the KL divergence between the joint distributions $q(\bold x_{0:T})$, $p_\theta(\bold x_{0:T})$ of these two chains. A simplified and reweighted version of this objective can be written as:
\begin{equation}
    \mathbb{E}_{t\sim\mathcal{U}[1,T],\bold x_0\sim q(\bold x_0), \boldsymbol\epsilon\sim\mathcal{N}(\bold 0,\bold I)}\left[\|\boldsymbol\epsilon-\boldsymbol\epsilon_\theta(\bold x_t,t)\|^2\right].
\end{equation}
After training the denoising network $\boldsymbol\epsilon_\theta$, samples can be generated from Gaussian noise through the reverse chain.

Similarly, the Conditional Diffusion Models are formulated by adding a condition $c$ to all the distributions in the deduction with an objective involving $c$:
\begin{equation}
    \mathbb{E}_{t\sim\mathcal{U}[1,T],\bold x_0,c\sim q(\bold x_0,c), \boldsymbol\epsilon\sim\mathcal{N}(\bold 0,\bold I)}\left[\|\boldsymbol\epsilon-\boldsymbol\epsilon_\theta(\bold x_t,t,c)\|^2\right].
\end{equation}

\subsection{3D Generation as Iterative View Sampling}\label{sec:theory}

Our assumption is that the distribution of 3D assets, denoted as $q_a(\bold x)$, is equivalent to the joint distribution of its corresponding multiview images. Specifically, given camera sequence $\{\boldsymbol\pi_0,\boldsymbol\pi_1,\cdots,\boldsymbol\pi_N\}$, we have 
\begin{equation}
    q_a(\bold x) = q_i(\Gamma(\bold x, \boldsymbol\pi_0),\Gamma(\bold x, \boldsymbol\pi_1),\cdots,\Gamma(\bold x, \boldsymbol\pi_N)),
\end{equation}
where $q_i$ is the distribution of images observed from 3D assets, and $\Gamma(\cdot,\cdot)$ is the 3D-2D rendering operator. This assumption is derived from the bijective correspondence between 3D assets and their multiview projections, given a sufficient number of views. The joint distribution can be factorized into a series of conditioned distributions:
\begin{equation}\label{eq:original}
\begin{split}
    q_a(\bold x)=\ &q_i(\Gamma(\bold x, \boldsymbol\pi_0))\cdot\\
    &q_i(\Gamma(\bold x, \boldsymbol\pi_1)|\Gamma(\bold x, \boldsymbol\pi_0))\cdot\\
    &\cdots\\
    &q_i(\Gamma(\bold x, \boldsymbol\pi_N)|\Gamma(\bold x, \boldsymbol\pi_0),\cdots,\Gamma(\bold x, \boldsymbol\pi_{N-1}))
\end{split}.
\end{equation}
It can be noticed that the conditional distributions exhibit an iterative arrangement. By sampling $\Gamma(\bold x, \boldsymbol\pi_n)$ step by step with previous samples as conditions, the joint multiview images are generated, thus directly determining the 3D asset.

In practice, however, multiview images are also difficult to obtain. To use unstructured 2D image collections, we construct training data using depth-based image warping. 
First, we substitute the original condition images in Eq.~\ref{eq:original}, \ie, \{$\Gamma(\bold x, \boldsymbol\pi_{k}),k=1,\ldots,n-1$\} for $\Gamma(\bold x, \boldsymbol\pi_{n})$, as $\Pi(\Gamma(\bold x, \boldsymbol\pi_{k}), \boldsymbol\pi_n)$, where $\Pi(\cdot, \cdot)$ denotes the depth-based image warping operation that warps an image to a given target view using depth. As a result, Eq.~\ref{eq:original} can be rewritten as

\begin{equation}\label{eq:modified}
\begin{split}
    q_a(\bold x)\approx\ &q_i(\Gamma(\bold x, \boldsymbol\pi_0))\cdot\\
    &q_i(\Gamma(\bold x, \boldsymbol\pi_1)|\Pi(\Gamma(\bold x, \boldsymbol\pi_0), \boldsymbol\pi_1))\cdot\\
    &\cdots\\
    &q_i(\Gamma(\bold x, \boldsymbol\pi_N)|\Pi(\Gamma(\bold x, \boldsymbol\pi_0), \boldsymbol\pi_N),\cdots)
\end{split}.
\end{equation}
Under this formulation, we further eliminate the requirement for actual multiview images $\Gamma(\bold x, \boldsymbol\pi_{k})$ by only warping $\Gamma(\bold x, \boldsymbol\pi_{n})$ itself back-and-forth. The details can be found in Sec.~\ref{sec:data}.

\emph{Note that} unlike some previous 3D-aware GANs~\cite{chan2021pi,deng2021gram,chan2022efficient, or2022stylesdf,gu2021stylenerf}, we model generic objects and scenes without pose label or any canonical pose definition. 
We directly regard the image distribution $q_d$ in the datasets as $q_i(\Gamma(\bold x, \boldsymbol\pi_0))$, \ie., the distribution of 3D assets' first partial view. All other views $\pi_1,\cdots,\pi_N$ are considered to be \emph{relative} to the first view.
This way, we formulate 3D-aware generation as an \emph{unconditional--conditional} image generation task, where an unconditional model is trained for $q_i(\Gamma(\bold x, \boldsymbol\pi_0))$ and a conditional model is trained for other terms $q_i(\Gamma(\bold x, \boldsymbol\pi_n)|\Pi(\Gamma(\bold x, \boldsymbol\pi_0), \boldsymbol\pi_n),\cdots)$.

\section{Approach}

\begin{figure*}[t]
	\centering
	\includegraphics[width=0.855\textwidth]{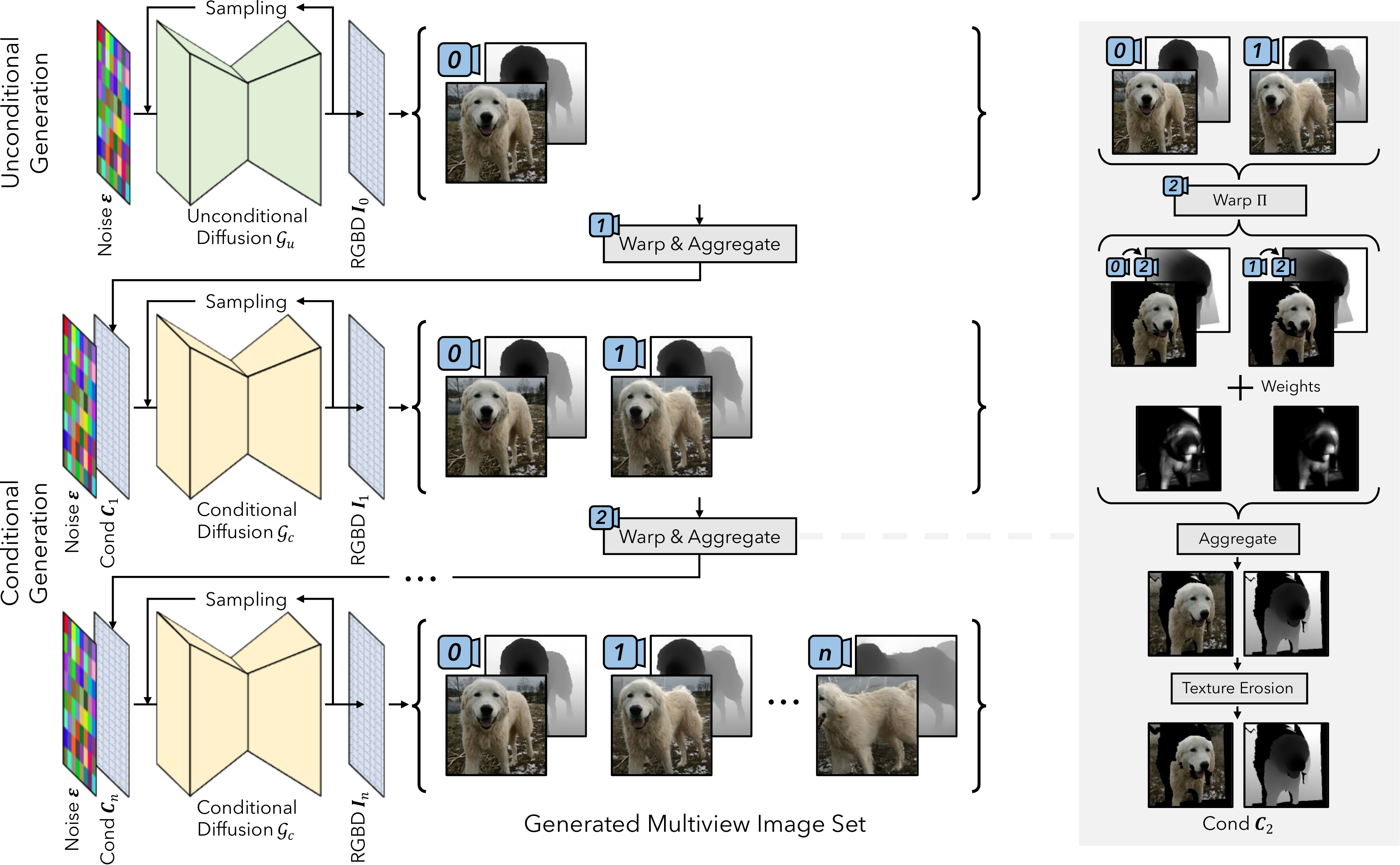}
    \vspace{-6pt}
	\caption{The overall framework. Our method contains two diffusion models $\mathcal{G}_u$ and $\mathcal{G}_c$. $\mathcal{G}_u$ is an unconditional model for randomly generating the first view, and $\mathcal{G}_c$ is a conditional generator for novel views. With \emph{aggregated} conditioning, multiview images are obtained iteratively by refining and completing previously synthesized views. For fast arbitrary-view synthesis, one can run 3D fusion or image-based rendering to synthesize new target views.}
	\label{fig:pipeline}
    \vspace{-5pt}
\end{figure*}

As per our problem formulation in Sec.~\ref{sec:theory}, our first step is to prepare the data, which includes the construction of RGBD images and the implementation of the warping algorithm (Sec.~\ref{sec:data}). 
We then train an unconditional RGBD diffusion model and a conditional model, parameterizing the unconditional term (the first one) and conditional terms (the others) in Eq.~\ref{eq:modified}, respectively (Sec.~\ref{sec:training}). After training, our method can generate diverse 3D-aware image samples with a broad camera pose range (Sec.~\ref{sec:inference}). The inference framework of our method is depicted in Fig.~\ref{fig:pipeline}.

\subsection{Data Preparation}\label{sec:data}

\paragraph{RGBD image construction}

To achieve RGBD warping, additional depth information is required for each image. We employ an off-the-shelf monocular depth estimator~\cite{ranftl2020towards} to predict depth map as it generalizes well to the targeted datasets with diverse objects and scenes.

\vspace{-10pt}
\paragraph{RGBD-warping operator}

The RGBD-warping operation $\Pi$ is a geometry-aware process determining the relevant information of partial RGBD observations under novel viewpoints. It takes a source RGBD image $\bold I_s=(\bold C_s, \bold D_s)$ and a target camera $\boldsymbol\pi_t$ as input, and outputs the visible image contents under target view $\bold I_t=(\bold C_t, \bold D_t)$ and a visibility mask $\bold M_t$, \ie, $\Pi: (\bold I_s, \boldsymbol\pi_t)\rightarrow(\bold I_t, \bold M_t).$
Our warping algorithm is implemented using a mesh-based representation and rasterizer. For an RGBD image, we construct a mesh by back-projecting the pixels to 3D vertices and defining edges for adjacent pixels on the image grid.

\begin{figure}[t]
	\centering
	\includegraphics[width=0.33\textwidth]{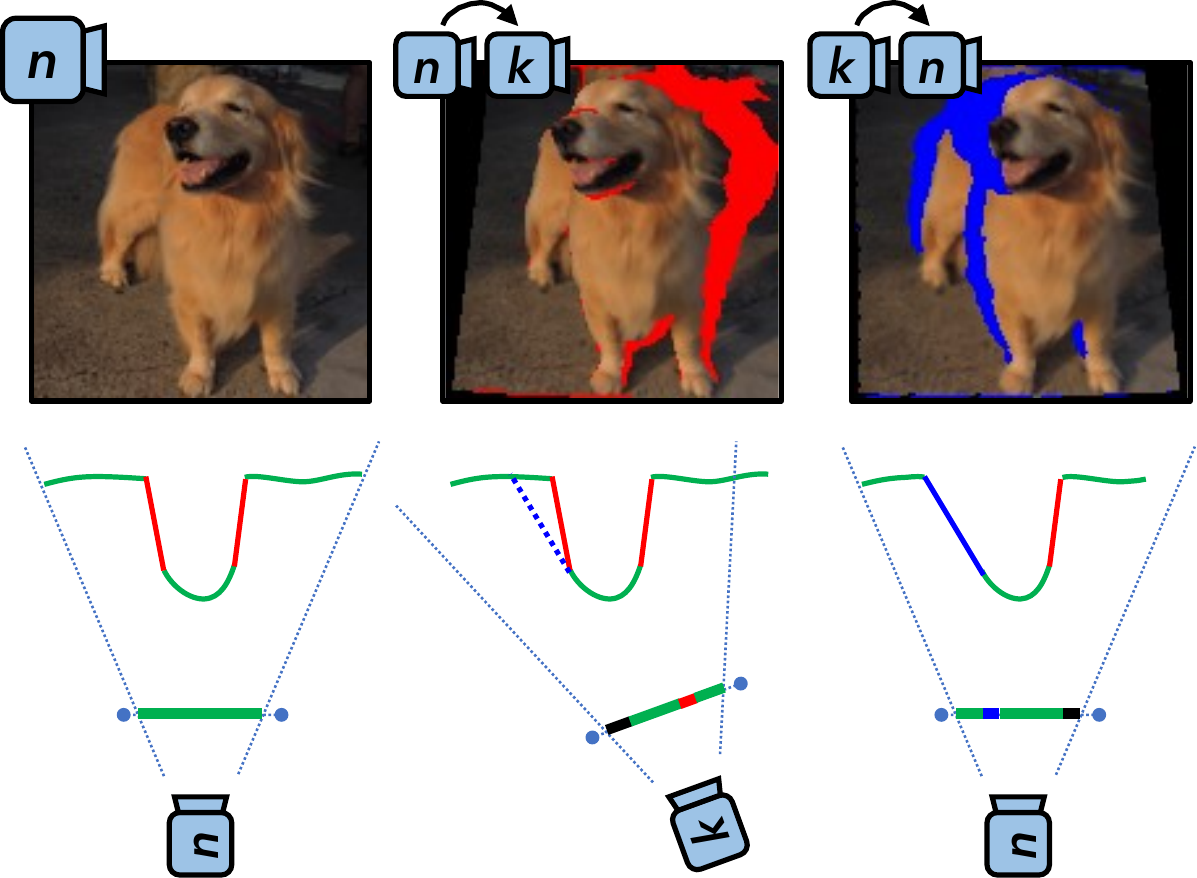}
	\vspace{-5pt}
	\caption{Illustration of forward-backward warpping.}
	\label{fig:warp}
    \vspace{-10pt}
\end{figure}

\vspace{-10pt}
\paragraph{Training pair construction}

To model the conditional distributions in Eq.~\ref{eq:modified}, data--condition pairs that comprise $\Gamma(\bold x, \boldsymbol\pi_{n})$ and $\Pi(\Gamma(\bold x, \boldsymbol\pi_{k}), \boldsymbol\pi_n)$ are required. 
Inspired by AdaMPI~\cite{han2022single}, we adopt a \emph{forward-backward warping} strategy to construct the training pairs from only $\Gamma(\bold x, \boldsymbol\pi_{n})$ without the need for actual images of $\Gamma(\bold x, \boldsymbol\pi_{k})$. Specifically, the target RGBD images are firstly warped to novel views and then warped back to the original target views. This strategy creates holes in the images which is caused by geometry occlusion. 
Despite its simplicity, conditions constructed using this strategy are equivalent to warp real images to the target views for Lambertian surfaces, or approximations of them for non-Lambertian regions:
\begin{equation}\label{eq:cond}
    \Pi(\Gamma(\bold x, \boldsymbol\pi_{k}), \boldsymbol\pi_n)\approx\Pi(\Pi(\Gamma(\bold x, \boldsymbol\pi_n), \boldsymbol\pi_{k}), \boldsymbol\pi_n).
\end{equation}
This is because the difference between  $\Gamma(\bold x, \boldsymbol\pi_{k})$ and $\Pi(\Gamma(\bold x, \boldsymbol\pi_n), \boldsymbol\pi_{k})$, \ie, the holes for scene contents not visible at view $\pi_{n}$, will be invisible again when wrapped back to $\pi_{n}$, and therefore become irrelevant. See Fig.~\ref{fig:warp} for an illustration of our training pair construction based on this forward-backward warping strategy. 

\subsection{Training}\label{sec:training}

\subsubsection{Unconditional RGBD generation}\label{sec:uncond}
We first train an unconditional diffusion model $\mathcal{G}_u$ to handle the distribution of all 2D RGBD images (the first term $q_i(\Gamma(\bold x, \boldsymbol\pi_0))$ in Eq.~\ref{eq:modified}). As mentioned, we directly regard the image distribution $q_d$ in the datasets as $q_i$, \ie., the distribution of 3D assets' partial observations, and train the diffusion model on the constructed RGBD images $\bold I\sim q_d(\bold I)$ to parameterize it.

We adopt the ADM network architecture from \cite{dhariwal2021diffusion} with minor modifications to incorporate the depth channel. For datasets with class labels (\eg, ImageNet~\cite{imagenet}), classifier-free guidance~\cite{ho2022classifier} is employed with a label dropping rate of $10\%$.

\subsubsection{Conditional RGBD completion and refining}

We then train a conditional RGBD diffusion model $\mathcal{G}_c$ for sequential view generation (the remaining terms $q_i(\Gamma(\bold x, \boldsymbol\pi_n)|\Pi(\Gamma(\bold x, \boldsymbol\pi_0), \boldsymbol\pi_n),\cdots)$  in Eq.~\ref{eq:modified}). The constructed data pairs $(\bold I$, $\Pi(\Pi(\bold I, \boldsymbol\pi_{k}), \boldsymbol\pi_n))$ using the forward-backward warping strategy are used to $\mathcal{G}_c$.
Instead of predefining the camera sequences $\{\boldsymbol\pi_n\}$ for training, we randomly sample relative camera pose from Gaussian distribution, which can make the process more flexible meanwhile keeping the generalization ability.

Our conditional models are fine-tuned from their unconditional counterparts. Specifically, we concatenate the additional condition, \ie, a warped RGBD image with mask, with the original noisy image to form the new network input. The holes on the condition RGBD image are filled with Gaussian noise. Necessary modifications to the network structure are made to the first layer to increase the number of input channels. Zero initialization is used for the added network parameters. Classifier-free guidance is not applied to these conditions.

We apply several data augmentation strategies to the constructed conditions for training. We found such augmentations can improve the performance and stability of the inference process.

\vspace{-10pt}
\paragraph{Blur augmentation}

\begin{figure}[t]
	\centering
	\includegraphics[width=0.47\textwidth]{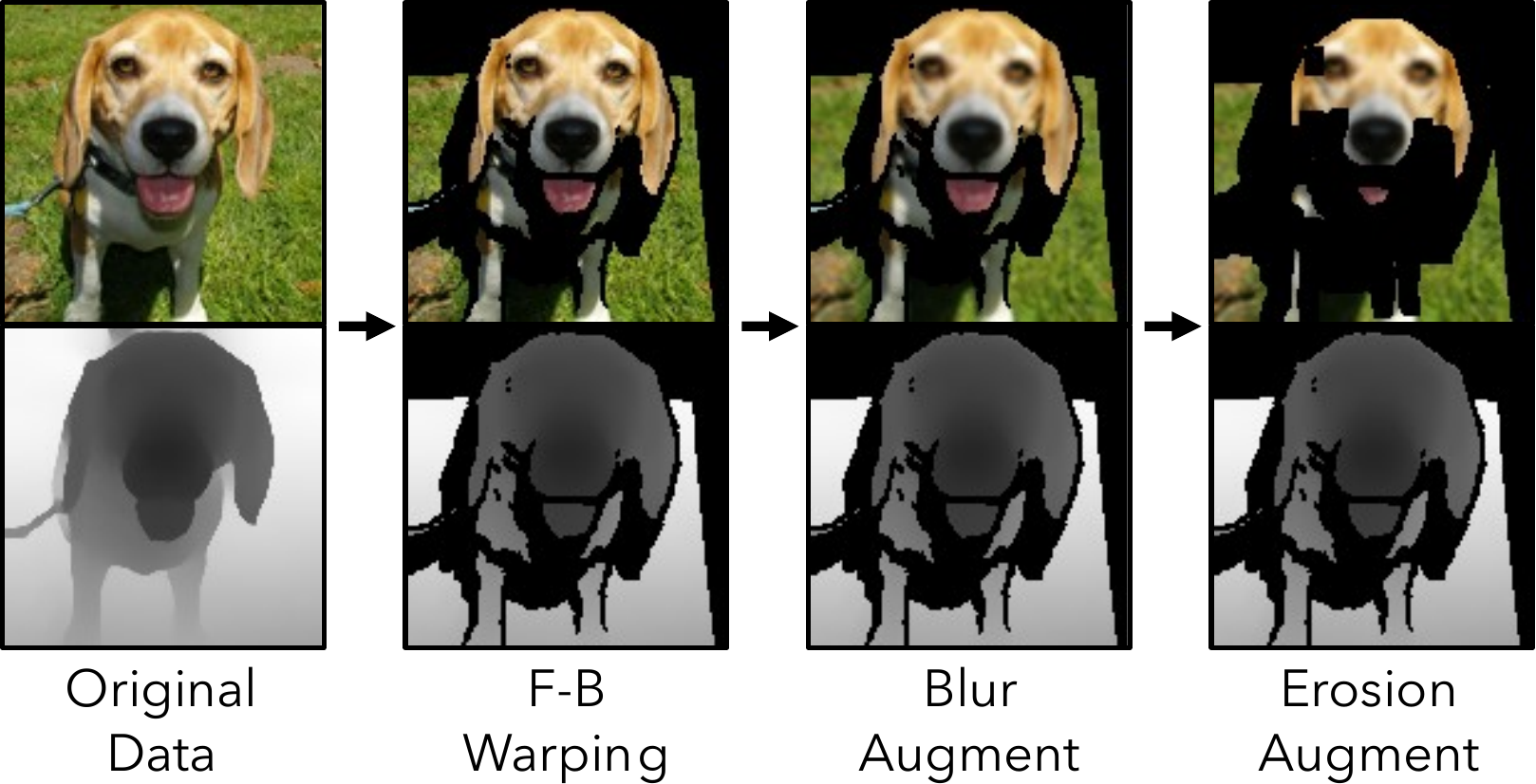}
	\vspace{-5pt}
	\caption{Illustration of our condition generation process.}
	\label{fig:data}
    \vspace{-6pt}
\end{figure}

The RGBD warping operation introduces image blur due to the low-pass filtering that occurs during interpolation and resampling in mesh rasterization. The forward-backward warping strategy involves two image warping steps, while only one is utilized during inference. To mitigate this gap, for the constructed conditions, we randomly replace the unmasked pixels in twice-warped images by pixels in the original images with a predefined probability and then apply Gaussian blur with random standard deviations (Fig.~\ref{fig:data}). This augmentation expands the training condition distribution to better reflect those encountered at inference time.

\vspace{-10pt}
\paragraph{Texture erosion augmentation}
The textures located close to depth discontinuities on the condition images have a negative impact on the image generation quality. This phenomenon can be attributed to two causes. Firstly, in-the-wild images contain complex view-dependent lighting effects, particularly near object boundaries (consider the Fresnel effect, rim light, subsurface scattering, \etc.).  These unique features serve as strong indicators of the edges of foreground objects, hindering the ability of the conditional model to generate appropriate geometry in novel views. Secondly, the estimated depth map is not perfect and may incur segmentation errors around object edges.
To address this issue, we perform random erosion on the texture component of the constructed conditions while leaving the depth unchanged (Fig.~\ref{fig:data}). This augmentation eliminates the problematic textural information near edges and leads to superior generation quality.

\subsection{Inference}\label{sec:inference}

With trained conditional and unconditional generative models, our 3D-aware iterative view sampling can be applied to obtain multiview images of a 3D asset:
\begin{equation}\label{eq:inference}
\begin{split}
    p_\theta(\bold I_0,\bold I_1,\cdots,\bold I_N)\approx\ &p_\theta(\bold I_0)\cdot\\
    &p_\theta(\bold I_1|\Pi(\bold I_0, \boldsymbol\pi_1))\cdot\\
    &\cdots\\
    &p_\theta(\bold I_N|\Pi(\bold I_0, \boldsymbol\pi_N),\cdots)
\end{split}.
\end{equation}
One can define a camera sequence that covers the desired views for multiview image synthesis. This camera sequence can be set arbitrarily to a large extent. Such flexibility is provided by random warping during the training stage. Following the given camera sequence, novel views are sampled one after the other iteratively, with all previously sampled images as conditions.

\vspace{-10pt}
\paragraph{Condition aggregation}
There remains a question of how our trained conditional diffusion models can be conditioned by all previously sampled images. We have tested both stochastic conditioning~\cite{watson2022novel, cai2022diffdreamer} and a new \emph{aggregated conditioning} strategy, and found the latter to be more effective for our task. As illustrated in Fig.~\ref{fig:pipeline} (right), aggregated conditioning collects information from previous images by performing a weighted sum across all warped versions of them:
\begin{equation}\begin{split}
    \bold C_n = \sum_{i=0}^{n-1}\bold W_{(i,n)}\Pi(\bold I_i, \boldsymbol\pi_n) \bigg/ \sum_{i=0}^{n-1}\bold W_{(i,n)},
\end{split}\end{equation}
where $\bold W_{(i,n)}$ is the weight map. The weight is  calculated for each pixel following the lumigraph rendering principles~\cite{buehler2001unstructured}. More details of the weight map computation can be found in Appendix~\ref{supp:cond}.

\paragraph{Fusion-based arbitrary-view synthesis}\label{sec:fusion}

In our original formulation, generating any novel view for an instance necessities running the diffusion model $\mathcal{G}_c$, which is inefficient for video generation and interactive applications. 

Here, we present a simple and efficient arbitrary-view generation solution based on fusing a fixed set of pre-generated views. Specifically, we first define a set of views uniformly covering the desired viewing range and generate images using the trained diffusion models. For any novel view, we warp the pre-generated views to it and aggregate them using a strategy following our condition aggregation (Eq.~\ref{eq:cond_aggregate} and \ref{eq:cond_aggregate_weight}). This strategy not only improves the speed for video generation, but also well preserves the texture detail consistency among different views.

\begin{figure*}[t!]
	\centering
	\includegraphics[width=1\textwidth]{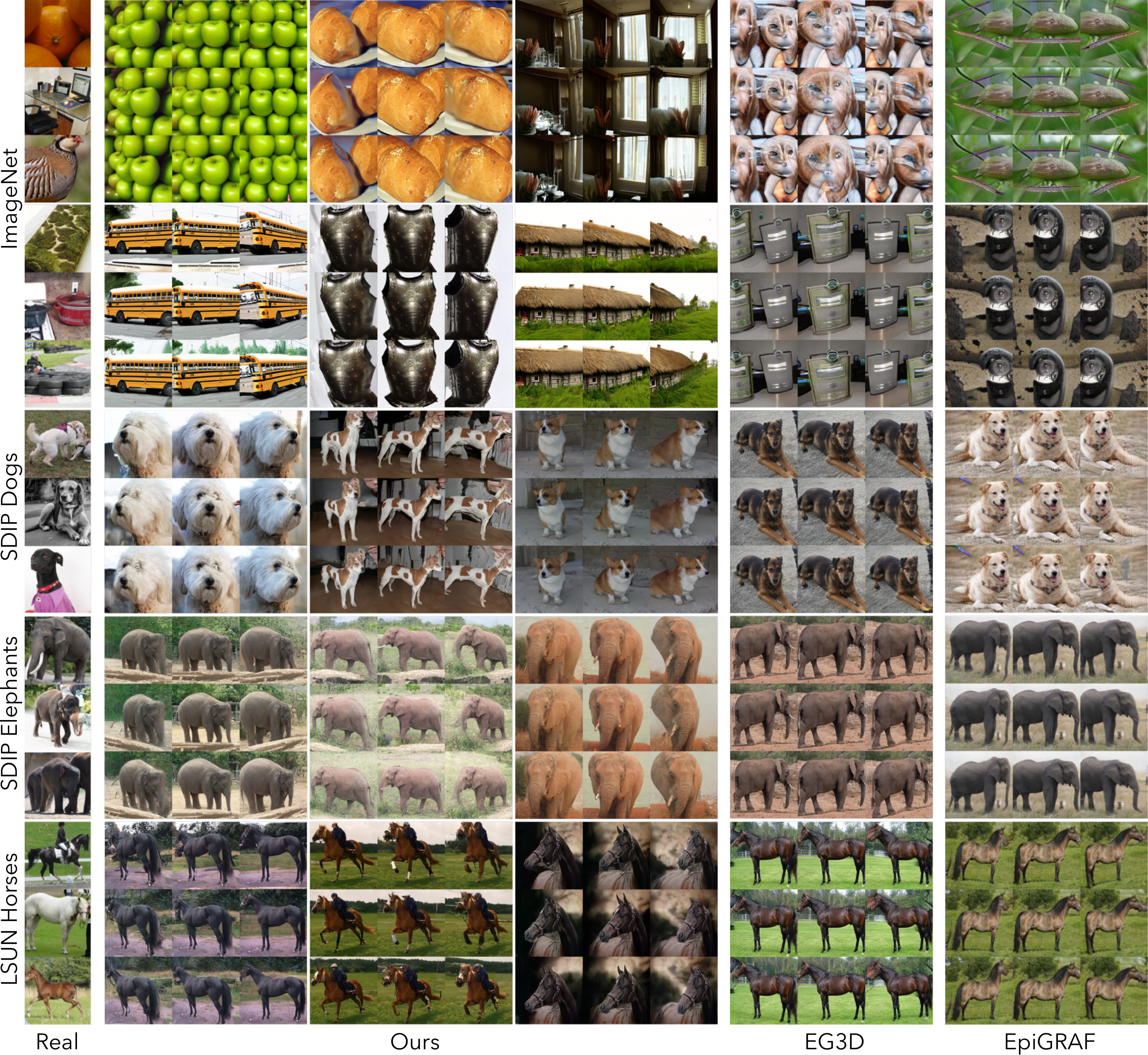}
	\vspace{-17pt}
	\caption{Multiview generation on ImageNet, SDIP Dogs, SDIP Elephants and LSUN horses datasets at $128^2$ resolution.}
	\label{fig:results}
    \vspace{-7pt}
\end{figure*}

\section{Experiments}

\paragraph{Implementation details}
We train our method on four datasets: ImageNet~\cite{imagenet}, SDIP Dogs~\cite{mokady2022self},
SDIP Elephants~\cite{mokady2022self} and LSUN Horses~\cite{yu2015lsun}. ImageNet is a large-scale dataset containing 1.3M images from 1000 classes. The other three are single-category datasets containing 125K, 38K, and 163K images, respectively. Images in these datasets are unaligned and contain complex geometry, which makes the 3D-aware image generation task challenging. We predict the depth maps using the MiDaS~\cite{ranftl2020towards} \emph{dpt\_beit\_large\_512} model.

Our experiments are primarily conducted on $128^2$ image resolution, and we will also demonstrate $256^2$ generation results using a diffusion-based supper-resolution model. We use the same network architecture and training setting as ADM~\cite{dhariwal2021diffusion} for the training on ImageNet and use a smaller version with channel halved on the other three datasets for efficiency. All our models are trained on 8 NVIDIA Tesla V100 GPUs with 32GB memory.
For ImageNet results, Classifier-free guidance weight of both unconditional and conditional networks is set to 3 for the shown samples and 0 for numerical evaluation.\footnote{All our code and trained models will be publicly released.}

\vspace{-10pt}
\paragraph{Inference speed} Evaluated on a NVIDIA Tesla V100 GPU, generating the initial view using $\mathcal{G}_u$ takes $20$s with 1000-step DDPM sampler, while generating one new view using $\mathcal{G}_c$ takes $1$s using 50-step DDIM sampler. 

\begin{figure}[t]
	\centering
	\includegraphics[width=0.47\textwidth]{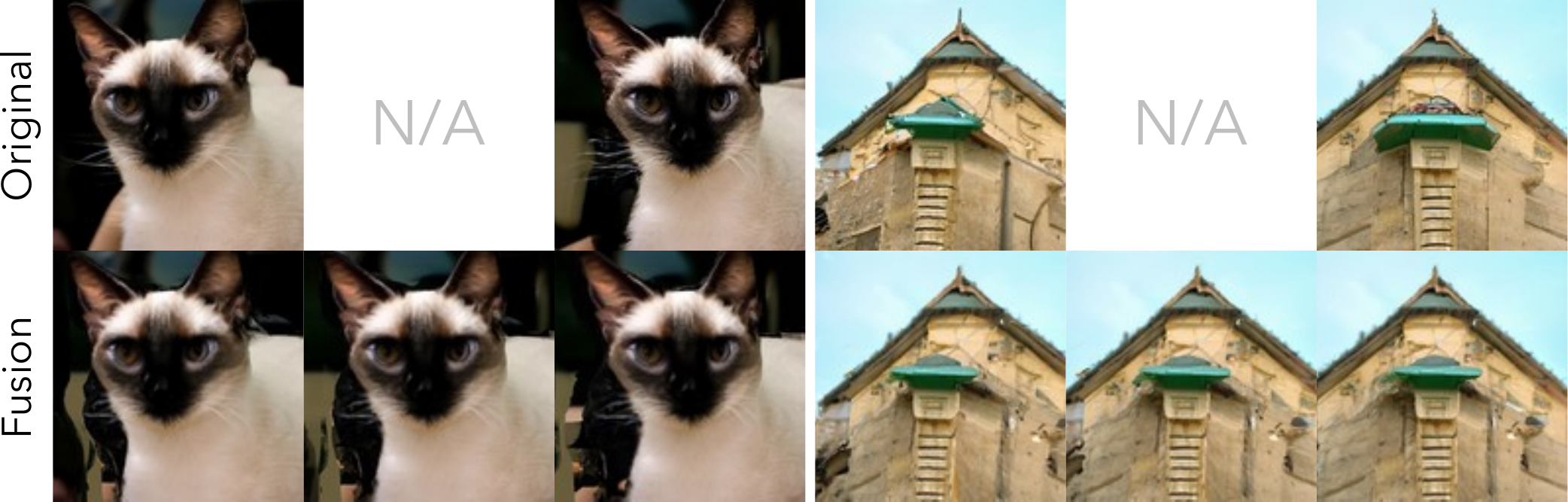}
	\vspace{-7pt}
	\caption{Images generated by our method with and without the fusion strategy.}
	\label{fig:fusion}
    \vspace{-3pt}
\end{figure}

\subsection{Visual Results}

Figure~\ref{fig:teaser}, \ref{fig:results} and \ref{fig:fusion} present some sampled multiview images from our method. As shown, our method can generate 3D-aware multiview images with diverse content and large view angle. High-quality 3D-aware images can be generated from in-the-wild image collections.

\subsection{Comparison to Prior Arts}

We compare our method with previous 3D-aware GANs including pi-GAN~\cite{chan2021pi}, EpiGRAF~\cite{skorokhodov2022epigraf} and EG3D~\cite{chan2022efficient}.\footnote{We are aware of a few recent works that train 3D-aware image generation on ImageNet, such as \cite{skorokhodov3d,sargent2023vq3d}. However, these works are concurrent to us, and their code or trained models are not publicly available.}  Since there is no pose label, the pose conditioning in EpiGRAF and EG3D are removed. Class labels are fed to the generator and discriminator instead. Note that no depth map is used by these methods.

For quantitative evaluation, we measure the Fréchet inception distance (FID)~\cite{heusel2017gans} and Inception Score (IS)~\cite{salimans2016improved} using 10K randomly generated samples and the whole real images set.
Following past practice \cite{chan2021pi}, camera poses are randomly sampled from Gaussian distributions with $\sigma=0.3$ and $0.15$ for the yaw and pitch angles, respectively. The results are shown in Table~\ref{tab:comparison_quality}. Some visual examples are presented in Fig.~\ref{fig:results}.

For the results on ImageNet, Table~\ref{tab:comparison_quality} shows that our results are significantly better than EpiGRAF and EG3D, while pi-GAN clearly underperformed. This large performance gain demonstrates the superior capability of our method for modeling diverse, large-scale image data. The visual examples also show the better quality of our results.

On other single-category datasets that have smaller scales, the quantitative results of the three methods are comparable: our method is slightly worse than EG3D and slightly better than EpiGRAF. However, their results often exhibit unrealistic 3D geometry. As can be observed from Fig.~\ref{fig:results}, both EG3D and EpiGRAF generated `planar' geometries and hence failed to produce the realistic 3D shapes of the synthesized objects, leading to wrong visual parallax when viewed with different angles.

\subsection{Large View Synthesis}\label{sec:large_view}
In this section, we further test the modeling capability of our conditional diffusion model $\mathcal{G}_c$, particularly under long camera trajectories for large view synthesis.

\vspace{-10pt}
\paragraph{Performance \emph{w.r.t.} view range} We first test our image generation quality under different view ranges. We define a long camera sequence $\{\bold \pi_n\}$ which forms a sampling grid with 9 columns for yaw and 3 rows for pitch, respectively. The resultant 27 views have angles ranging $\pm0.6$ for yaw (\ie, $\sim\!70^\circ$ range) and $\pm0.15$ for pitch (\ie, $\sim\!17^\circ$ range), respectively. The numerical results in Table~\ref{tab:large_view} show that the quality degrades moderately as the view range gets larger. The quality drop can be attributed to two reasons: domain drifting and data bias (see Appendix~\ref{supp:discussion} for discussions).
Figure~\ref{fig:large_view} shows all 27 views of two samples. The visual quality for large angles remains reasonable.

\begin{figure}[t]
	\centering
	\includegraphics[width=0.47\textwidth]{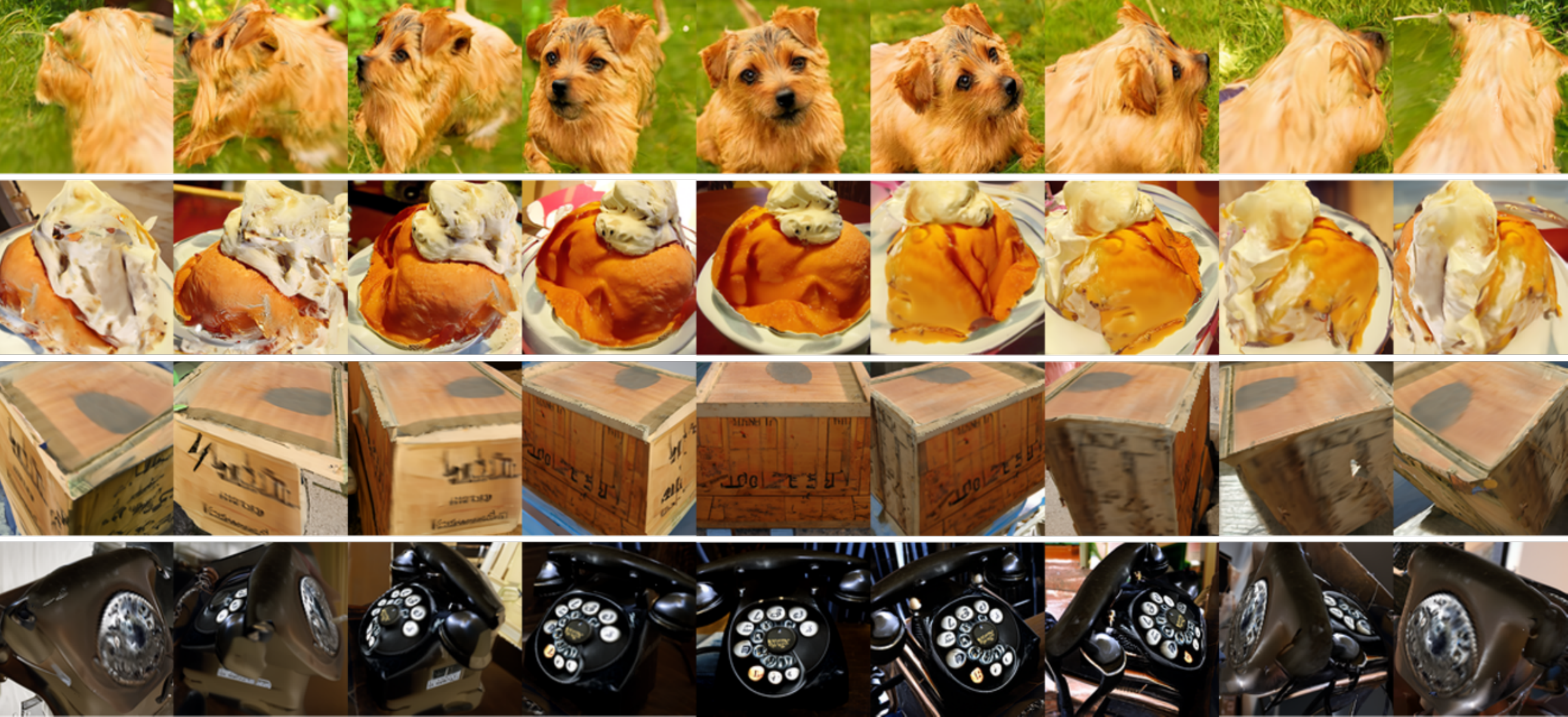}
	\vspace{-5pt}
	\caption{Curated 360$^\circ$ generation results on ImageNet.}
	\label{fig:surround}
    \vspace{0pt}
\end{figure}

\vspace{-10pt}
\paragraph{360$^\circ$ generation} 
We conducted an evaluation of 360$^\circ$ generation on ImageNet and found that our approach demonstrates efficacy in certain scenarios, as shown in Fig.~\ref{fig:teaser} and \ref{fig:surround}. 
Note that 360$^\circ$ generation of unbounded real-world scenes is a challenging task. One significant contributor to this challenge is the data bias problem, where rear views of objects are frequently underrepresented. 

\setlength{\arrayrulewidth}{0.5mm}
\setlength{\tabcolsep}{3pt}
\renewcommand{\arraystretch}{1.0}
\begin{table}
	\centering
	\caption{Quantitative comparison of generation quality with $\text{FID}$ and IS scores using 10K generated samples.}
	\label{tab:comparison_quality}
	\centering
	\small  
	\vspace{-8pt}
	\begin{tabular}{cccccc}
		\toprule
		      & \multicolumn{2}{c}{ImageNet} & ~Dog & \!Elephant\! & Horse \\
		Method     & ~~FID$\downarrow$~~ & ~~~IS$\uparrow$~~~ & ~FID$\downarrow$ & FID$\downarrow$ & FID$\downarrow$ \\
		\midrule
        pi-GAN~\cite{chan2021pi} & 138 & 6.82 & 115 & 71.0 & 92.6 \\
        EpiGRAF~\cite{skorokhodov2022epigraf} & 67.3 & 12.7 & 17.3 & 7.25 & 5.82 \\
		EG3D~\cite{chan2022efficient} & 40.4 & 16.9 & 9.83 & 3.15 & 2.61 \\
        \emph{Ours} & 9.45 & 68.7 & 12.0 & 6.00 & 4.01 \\
        \emph{Ours-fusion} & 14.1 & 61.4 & 14.7 & 11.0 & 10.2 \\
		\bottomrule
	\end{tabular}
	\vspace{0pt}
\end{table}

\setlength{\arrayrulewidth}{0.5mm}
\setlength{\tabcolsep}{3pt}
\renewcommand{\arraystretch}{1.0}
\begin{table}
	\centering
	\caption{Generation quality with various view ranges, measured with $\text{FID}$ and IS scores of 10K generated samples.
    }
	\label{tab:large_view}
	\centering
	\small  
	\vspace{-8pt}
	\begin{tabular}{lccccc}
		\toprule
		      & \multicolumn{2}{c}{ImageNet} & Dog & \!Elephant\! & Horse \\
		(\#views, yaw range)\!\!     & ~~FID$\downarrow$~ & ~~~IS$\uparrow$~~~~ & FID$\downarrow$ & FID$\downarrow$ & FID$\downarrow$ \\
		\midrule
        (\ \ 1, \ \ $0^\circ$) -- {\scriptsize $\mathcal{G}_u$\!\,only} & 7.85 & 85.2 & 8.48 & 4.06 & 2.50 \\
        (\ \ 9, $17^\circ$) & 8.90 & 74.9 & 11.5 & 6.22 & 3.52 \\
        (15, $35^\circ$) & 9.82 & 71.0 & 13.0 & 7.95 & 4.85 \\
        (21, $50^\circ$) & 11.2 & 66.1 & 14.9 & 10.1 & 6.75 \\
        (27, $70^\circ$) & 13.0 & 60.3 & 17.0 & 12.8 & 9.41 \\
		\bottomrule
	\end{tabular}
	\vspace{5pt}
\end{table}

\begin{figure}[t]
	\centering
	\includegraphics[width=0.47\textwidth]{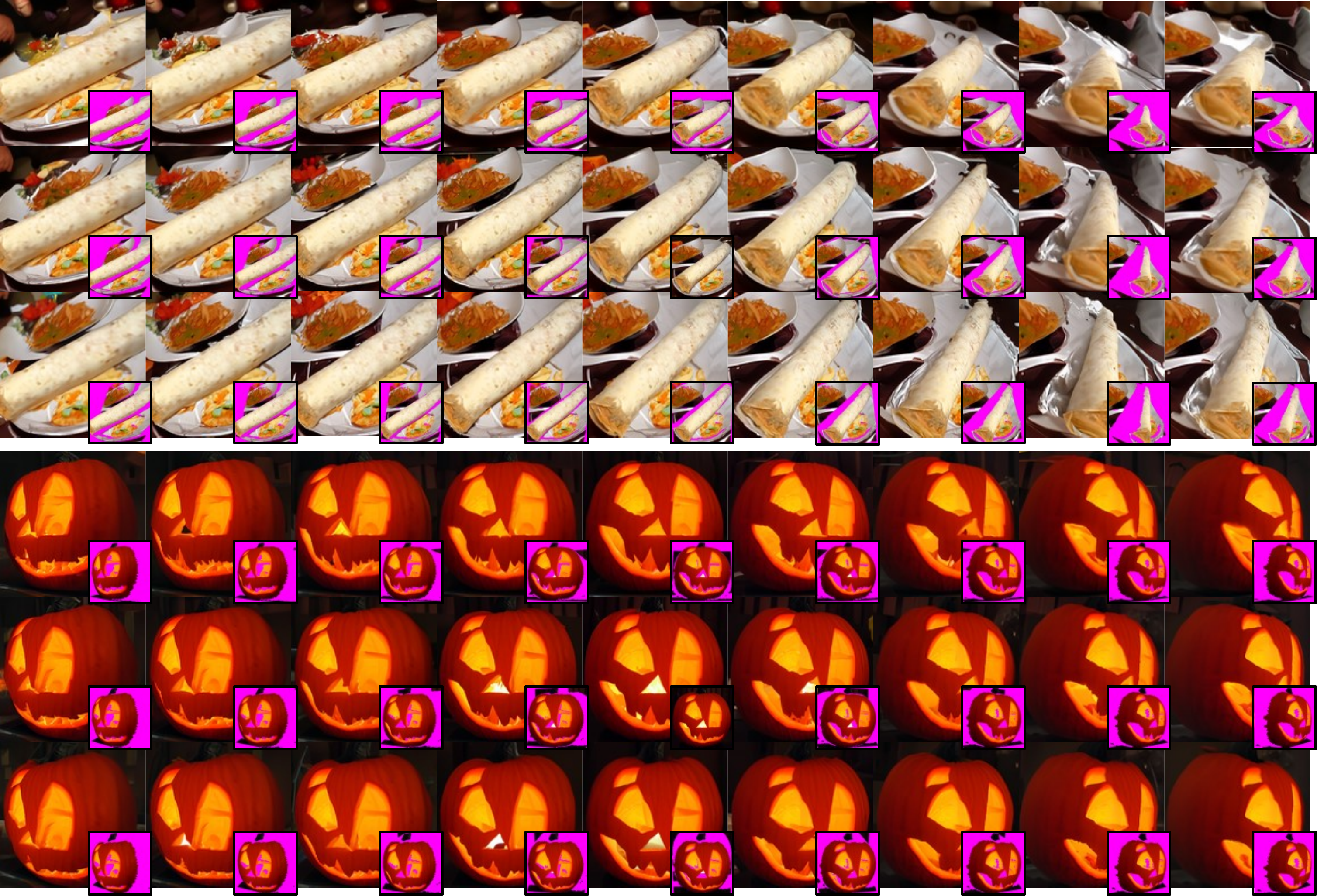}
	\vspace{-5pt}
	\caption{Large view synthesis results. To highlight the contribution of the conditional generator $\mathcal{G}_c$, we show a smaller figure where the regions invisible in the first view are masked with pink color.}
	\label{fig:large_view}
    \vspace{0pt}
\end{figure}

\subsection{Ablation Study}
\begin{figure}[t]
	\centering
	\includegraphics[width=0.47\textwidth]{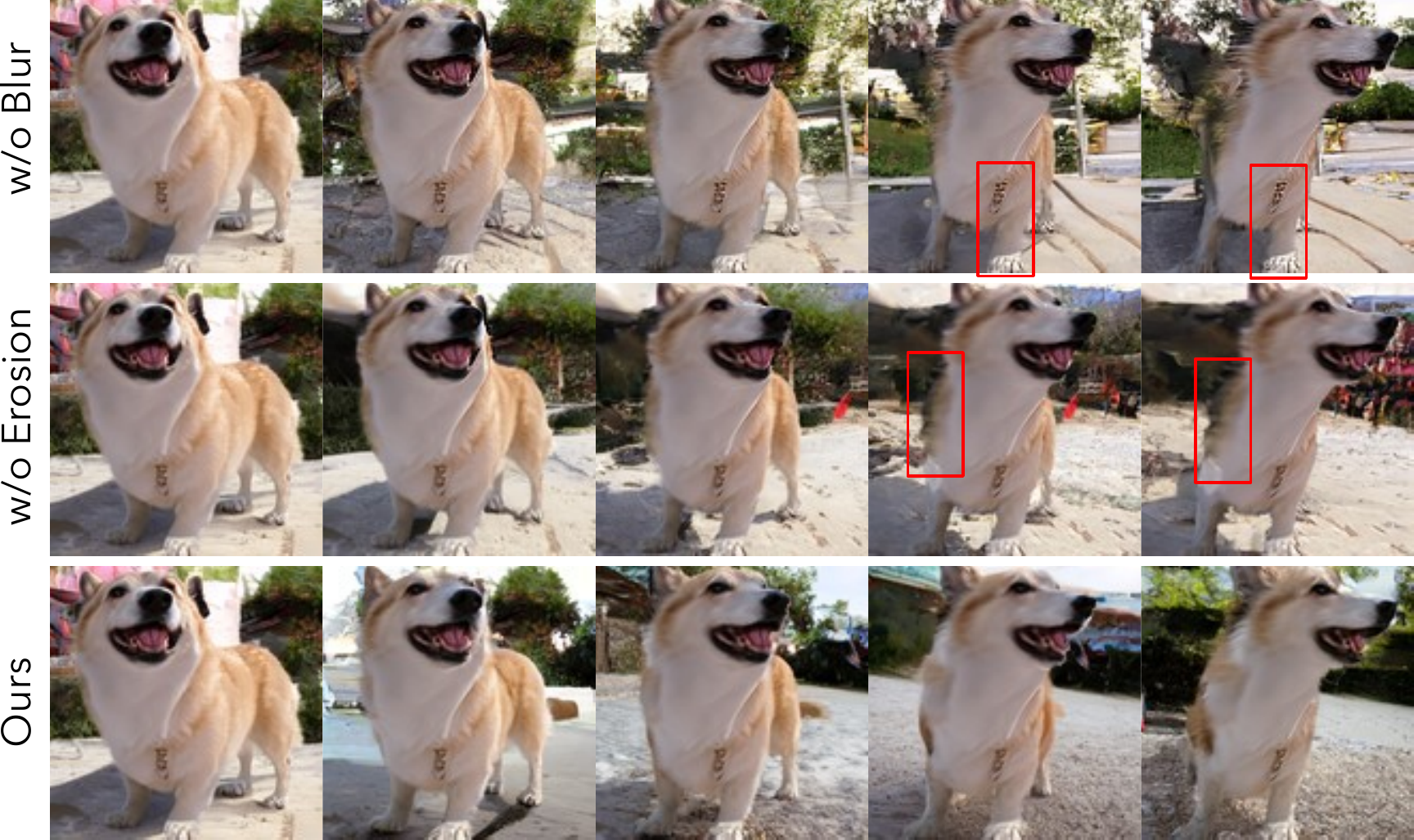}
	\vspace{-7pt}
	\caption{Ablation study on our proposed data augmentation strategies. Noticeable artifacts are marked with box. }
	\label{fig:ablation}
    \vspace{-4pt}
\end{figure}

\paragraph{Data augmentation strategies} We train two conditional models on the SDIP Dog dataset without augmentation and compare the results both visually and quantitatively to verify their effectiveness. 27 views are synthesized for each generated instance following the evaluation in Sec.~\ref{sec:large_view}.
Figure~\ref{fig:ablation} shows that without blur augmentation, the generated images become excessively sharp after a short view sampling chain, which is also detrimental in terms of the FID metric (Table~\ref{tab:ablation}). Additionally, without texture erosion augmentation, unreliable information on the edges of the depth map can negatively impact the conditional view sampling process, resulting in poor large-view results. This decrease in quality is also evident in the FID metrics. With all of our proposed augmentations enabled, we achieve the best results both visually and quantitatively.

\setlength{\arrayrulewidth}{0.5mm}
\setlength{\tabcolsep}{8pt}
\renewcommand{\arraystretch}{1.0}
\begin{table}
	\centering
	\caption{Ablation study on the proposed condition augmentation strategies. The FID-2K metric on the SDIP Dog dataset are reported.}
	\label{tab:ablation}
	\centering
	\small  
	\vspace{-8pt}
	\begin{tabular}{cccc}
		\toprule
		   (\#views, yaw range)\!\!  & \emph{Ours} & \!\!w/o erosion\!\! & \!\!w/o blur \\
		\midrule
        (\ \ 9, $17^\circ$) & 18.1 & 18.4 & 19.1 \\
        (15, $35^\circ$) & 19.2 & 20.9 & 22.2 \\
        (27, $70^\circ$) & 23.1 & 26.8 & 31.6 \\
		\bottomrule
	\end{tabular}
	\vspace{4pt}
\end{table}

\vspace{-10pt}
\paragraph{Fusion-based view synthesis} Table~\ref{tab:comparison_quality} shows the quantitative results of our efficient, fusion-based arbitrary-view synthesis solution. For this solution, we first generate 27 fixed views with $70^\circ$ yaw range and $17^\circ$ pitch range (Sec.~\ref{sec:large_view}) and use them to generate novel views. After generating these 27 views, it can run at 16 fps to generate arbitary novel views with our unoptimized mesh rendering implementation. Its FID score is still significantly lower than previous methods on ImageNet, but slightly higher compared to our original method. This is expected as the image-based fusion inevitably introduces blur and other distortions. Figure~\ref{fig:fusion} compares the image samples generated by our method with and without the fusion strategy. The results with smoothly-changing views in our \emph{supplementary video} are also generated with this fusion strategy. For the 360$^\circ$ renderings in the video, the results are obtained by fusing 15 views covering the upper hemisphere of camera viewpoints.

\begin{figure}[t]
	\centering
	\includegraphics[width=0.47\textwidth]{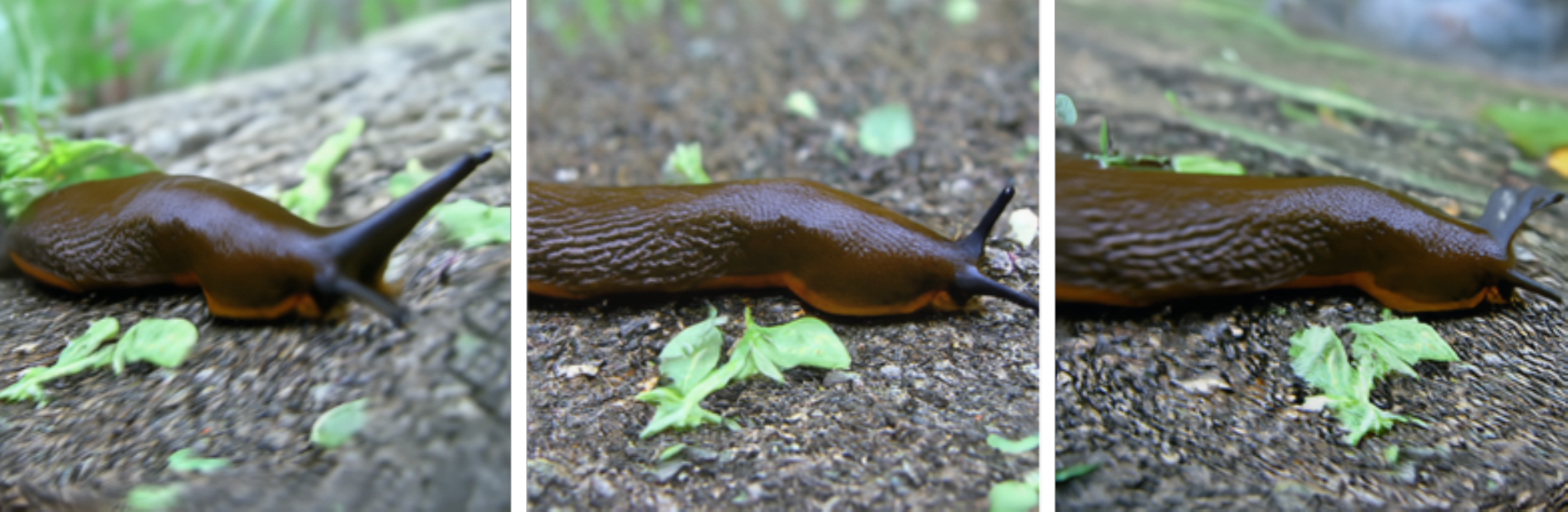}
	\vspace{-7pt}
	\caption{$256^2$ generation result upsampled from $128^2$ using diffusion-based super-resolution model.}
	\label{fig:superres}
    \vspace{-3pt}
\end{figure}

\subsection{Higher-Resolution Generation}
In theory, our method can be directly applied to train on higher-resolution images given sufficient computational resources. An efficient alternative is to apply image-space upsampling, which has been seen frequent use in previous 3D-aware GANs~\cite{chan2022efficient, or2022stylesdf, gu2021stylenerf}. We have implemented a $256^2$ DM conditioned on low-resolution images for image upsampling following Cascaded Diffusion~\cite{ho2022cascaded}. This model is trained efficiently by fine-tuning a pretrained $128^2$ unconditional model. Figure~\ref{fig:superres} shows one $256^2$ sample from this model; more can be found in Appendix~\ref{supp:results}.

\section{Conclusion}\label{sec:conclusion}

We have presented a novel method for 3D-aware image generative modelling. Our method is derived from a new formulation of this task: sequential unconditional-conditional generation of multiview images. We incorporate depth information to construct our training data using only still images, and train diffusion models for multiview image modeling. The training results on both large-scale multi-class dataset (\ie, ImageNet) and complex single-category datasets have collectively demonstrated the strong generative modelling power of our proposed method. 

{\small
\bibliographystyle{ieee_fullname}
\bibliography{ref}
}

\clearpage

\appendix
\section{More Training Details}

The network architecture of our method is adopted from ADM~\cite{dhariwal2021diffusion} with necessary modifications to support our task. Specifically, for the unconditional diffusion model $\mathcal{G}_u$, the first convolution layer's input channel and the last convolution layer's output channel are enlarged from 3 to 4 to incorporate the depth map. For the conditional diffusion model $\mathcal{G}_c$, the fist layer is enlarged to 10 channels to incorporate additional conditions (4 channels for the noisy RGBD image, 3 for the warped texture, 1 for the warped depth, 1 for the texture mask, and 1 for the depth ). 
For super-resolution model $\mathcal{G}_{sr}$, the fist layer is enlarged to 8 channels to include the low-resolution image as condition.
The parameters added for injecting the conditions are initialized as zero to remove the impact of conditions at the beginning of conditional model fine-tuning. 

All our models are trained with FP16 precision using 8 NVIDIA Tesla V100 GPU with 32GB memory FP16 training. We use the Adam~\cite{kingma2015adam} optimizer with a learning rate of $1e-4$ and $\beta$ set to $(0.9, 0.999)$ for all the datasets. The batchsize is set to 64. Exponential Moving Average (EMA) is enabled with smoothing rate of 0.9999 to boost the performance. More information about the networks and training process can be found in Table~\ref{tab:model}.

\setlength{\arrayrulewidth}{0.5mm}
\setlength{\tabcolsep}{8pt}
\renewcommand{\arraystretch}{1.0}
\begin{table}
	\centering
	\caption{Network and training details}
	\label{tab:model}
	\centering
	\small  
	\vspace{-9pt}
	\begin{tabular}{cccc}
		\toprule
		  Network & \!\!\#Params\!\! & Steps & \!\!\!Training time\!\!\! \\
		\midrule
		  ImageNet $\mathcal{G}_u$ & 422M & 1M & $\sim$ 11.1 Days \\
		  ImageNet $\mathcal{G}_c$ & 422M & 500K & $\sim$ 5.73 Days \\
		  ImageNet $\mathcal{G}_{sr}$ & 105M & 200K & $\sim$ 0.54 Days \\
            SDIP Dogs $\mathcal{G}_u$ & 105M & 500K & $\sim$ 1.29 Days \\
            SDIP Dogs $\mathcal{G}_c$ & 105M & 300K & $\sim$ 0.82 Days\\
            SDIP Dogs $\mathcal{G}_{sr}$ & 105M & 200K & $\sim$ 0.54 Days \\
            SDIP Elephants $\mathcal{G}_u$ & 105M & 300K & $\sim$ 0.77 Days\\
            SDIP Elephants $\mathcal{G}_c$ & 105M & 300K & $\sim$ 0.82 Days\\
            SDIP Elephants $\mathcal{G}_{sr}$ & 105M & 200K & $\sim$ 0.54 Days \\
            LSUN Horses $\mathcal{G}_u$ & 105M & 300K & $\sim$ 0.77 Days\\
            LSUN Horses $\mathcal{G}_c$ & 105M & 300K & $\sim$ 0.82 Days\\
            LSUN Horses $\mathcal{G}_{sr}$ & 105M & 200K & $\sim$ 0.54 Days \\
		\bottomrule
	\end{tabular}
	\vspace{-4pt}
\end{table}

\section{More Implementation Details}
\subsection{Condition Aggregation Details}\label{supp:cond}

In Sec.~4.3 of the main paper, we introduced our \emph{aggregated conditioning} strategy and here we present more details regarding aggregation weight computation.

To sample from $q_i(\Gamma(\mathbf{x}, \boldsymbol\pi_n)|\Pi(\Gamma(\bold x, \boldsymbol\pi_0), \boldsymbol\pi_n),\cdots)$, our condition aggregation collects information from previous images by performing a weighted sum across all warped versions of them:
\begin{equation}\begin{split}
    \bold C_n = \sum_{i=0}^{n-1}\bold W_{(i,n)}\Pi(\bold I_i, \boldsymbol\pi_n) \bigg/ \sum_{i=0}^{n-1}\bold W_{(i,n)},\label{eq:cond_aggregate}
\end{split}\end{equation}
where $\bold W_{(i,n)}$ is the weight map. The weight is calculated
for each pixel $(x,y)$. Let  $\bold p^{x,y}$ and $\bold n^{x,y}$ be the pixel's spatial position and normal in the world coordinate space and $d^{x,y}$ be its distance to the nearest masked pixel on image plane, we set
\begin{equation}
    \bold W_{(i,n)}^{x,y} = \phi\big(\frac{\boldsymbol o_i-\boldsymbol p^{x,y}}{\|\boldsymbol o_i-\boldsymbol p^{x,y}\|} \cdot\bold n^{x,y}\big) \psi(d^{x,y}), \label{eq:cond_aggregate_weight}
\end{equation}
where $\boldsymbol o_i$ is the camera center of the $i$-th view. $\phi(\cdot)$ and $\psi(\cdot)$ are scalar functions for balancing the weights of the two terms, which are empirically set as $\phi(a)=\exp(-20*\arccos(a))$ and $\psi(b)=b$.
The first term assign each pixel the least distorted information~\cite{buehler2001unstructured} and the second term suppresses the contribution of pixels near occlusion boundaries.

\section{More Experimental Results}\label{supp:results}

\subsection{More Ablation Study} 
\begin{figure}[t]
	\centering
	\includegraphics[width=0.35\textwidth]{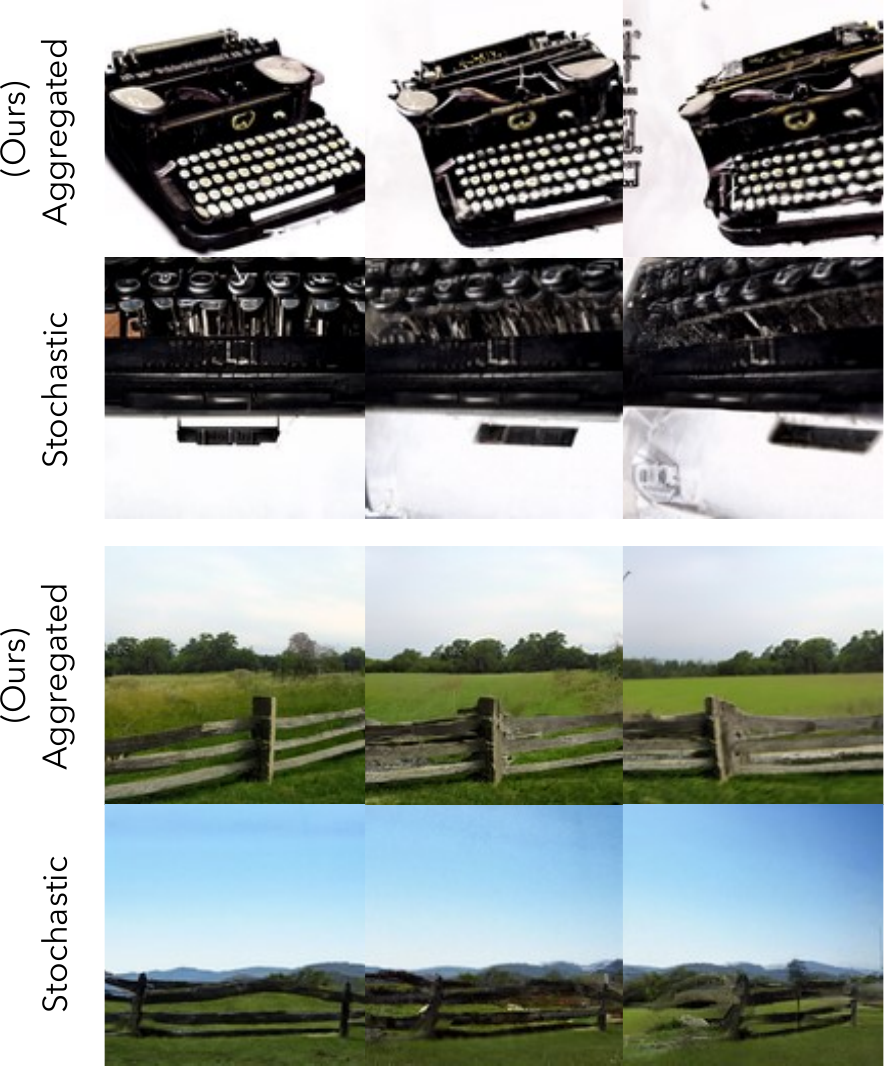}
    \vspace{-6pt}
	\caption{Comparison between stochastic conditioning and our aggregated conditioning.}
	\label{fig:stoch_cond}
    \vspace{-5pt}
\end{figure}

\paragraph{Multiview conditioning strategy}
We further compare the effectiveness of our \emph{aggregated conditioning} strategy against stochastic conditioning~\cite{watson2022novel, cai2022diffdreamer} in Fig.~\ref{fig:stoch_cond}.
For our task, stochastic conditioning is not suitable as it does not properly consider all previously-generated contents and will lead to inconsistency among different views.

\subsection{More Visual Results}

\begin{figure*}[t]
	\centering
	\includegraphics[width=0.825\textwidth]{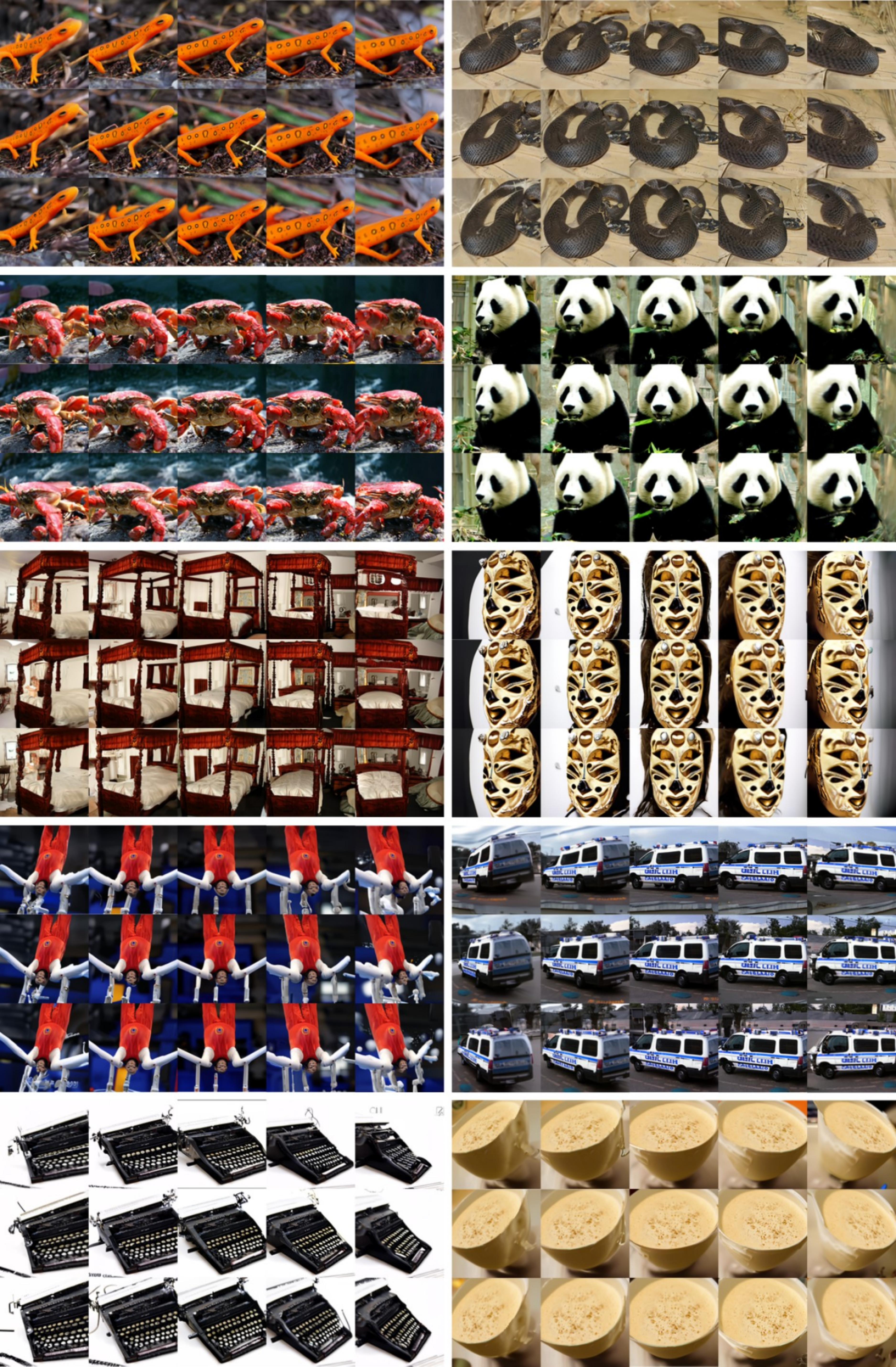}
    \vspace{-6pt}
	\caption{More $128^2$ multiview results on ImageNet.}
	\label{fig:more_results_1}
    \vspace{-5pt}
\end{figure*}

\begin{figure*}[t]
	\centering
	\includegraphics[width=0.825\textwidth]{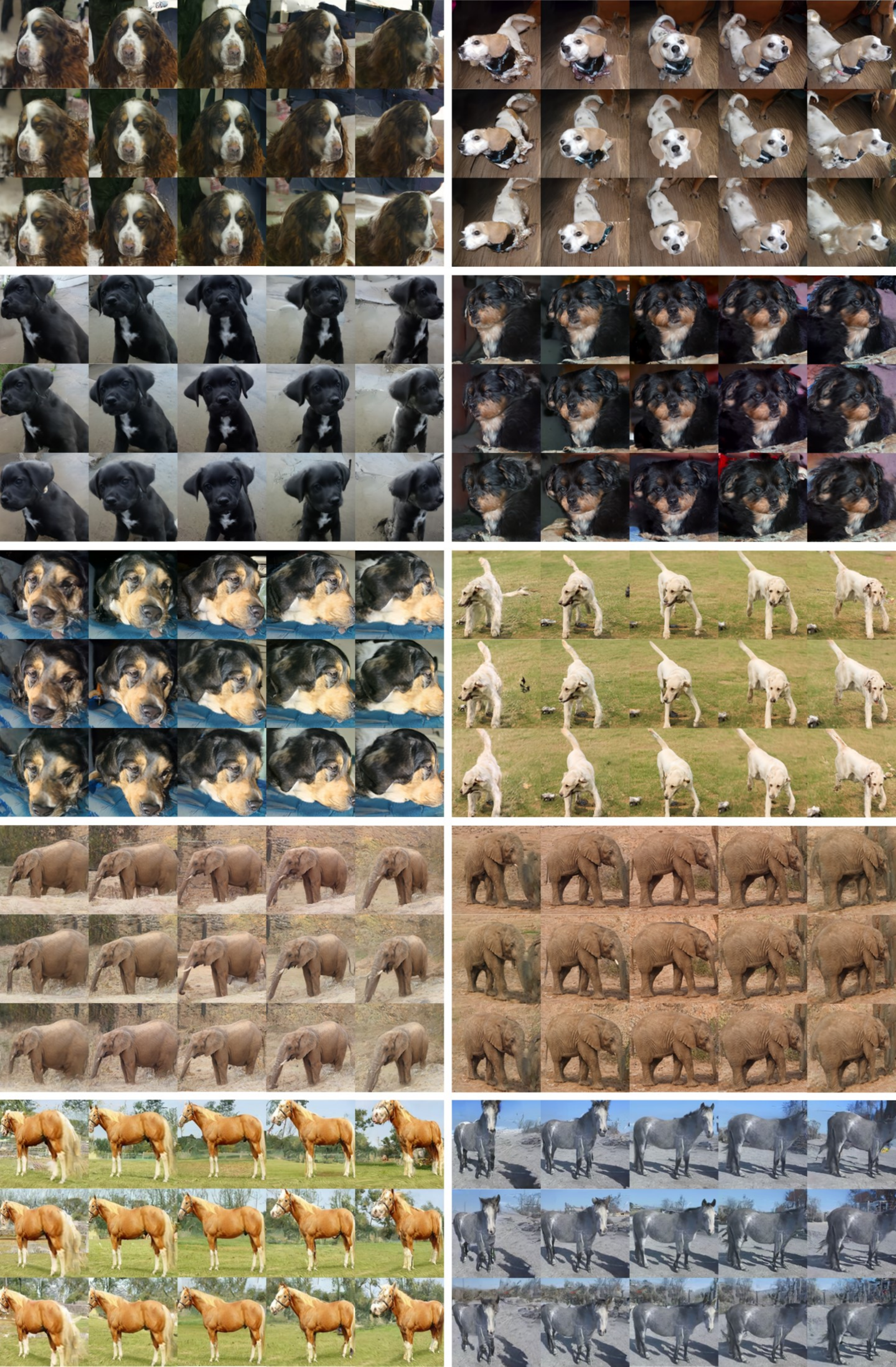}
    \vspace{-6pt}
	\caption{More $128^2$ multiview results for SDIP Dogs, SDIP Elephants, and LSUN Horses.}
	\label{fig:more_results_2}
    \vspace{-5pt}
\end{figure*}

\begin{figure*}[t]
	\centering
	\includegraphics[width=0.825\textwidth]{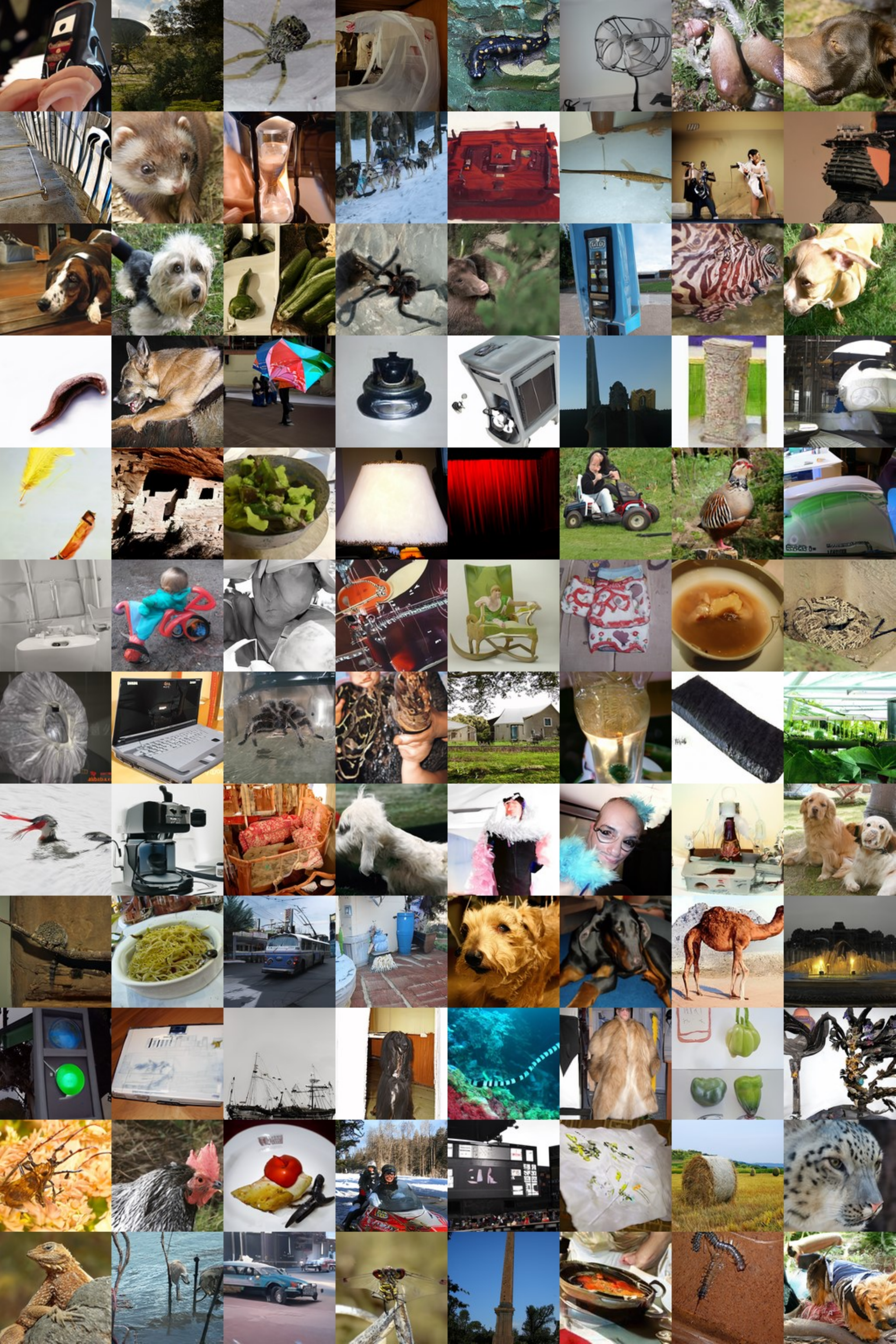}
    \vspace{-6pt}
	\caption{More uncurated $128^2$ single-view results for ImageNet. Note that the views are randomly sampled.}
	\label{fig:more_results_3}
    \vspace{-5pt}
\end{figure*}

\begin{figure*}[t]
	\centering
	\includegraphics[width=0.825\textwidth]{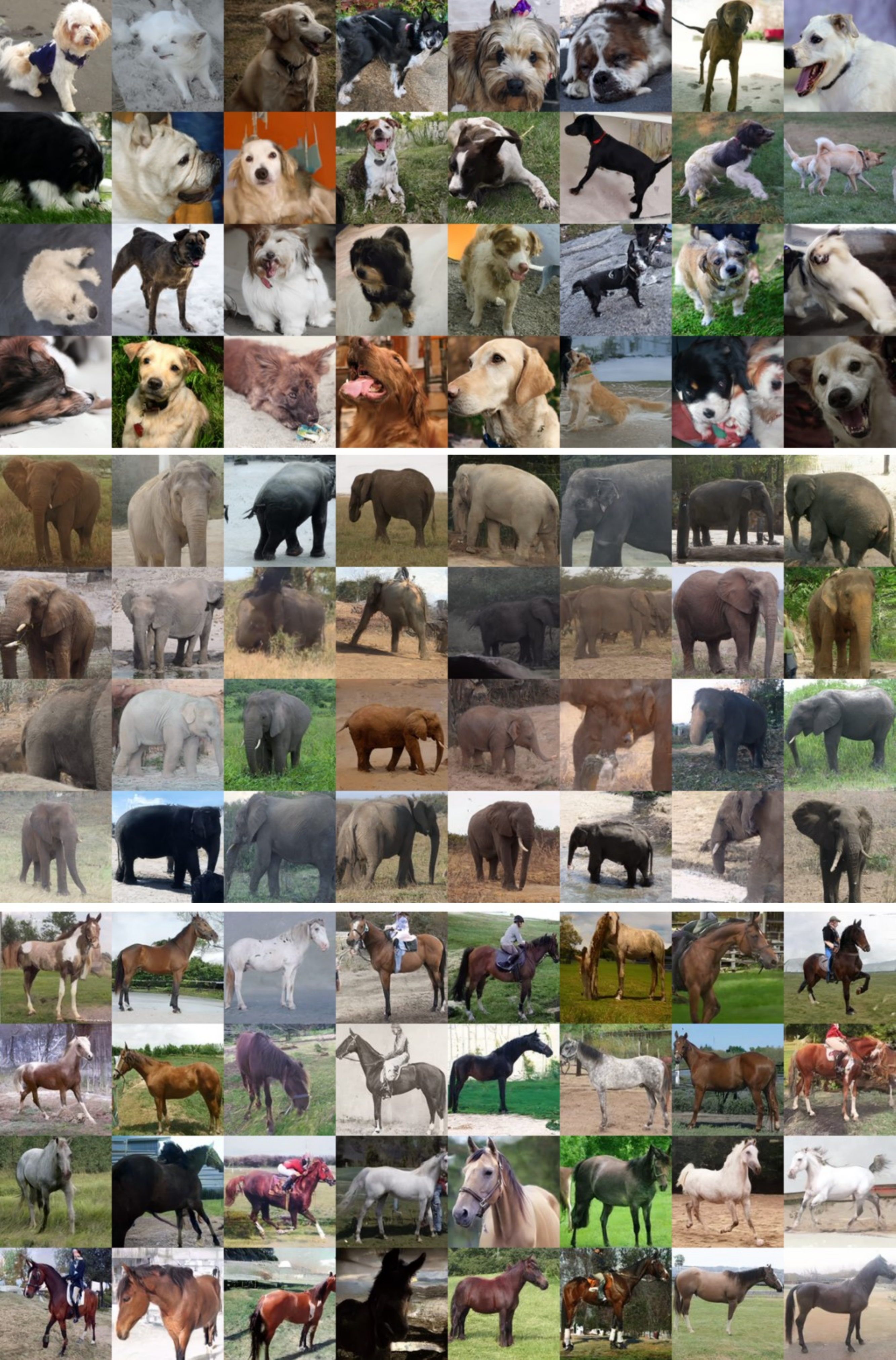}
    \vspace{-6pt}
	\caption{More uncurated $128^2$ single-view results for SDIP Dogs, SDIP Elephants, and LSUN Horses. Note that the views are randomly sampled.}
	\label{fig:more_results_4}
    \vspace{-5pt}
\end{figure*}

In Fig.~\ref{fig:more_results_1} and Fig.~\ref{fig:more_results_2}, we show more multiview results of $128^2$ resolution for each of the four datasets we tested. More uncurated results with camera pose randomly drawn from Gaussian distribution ($\sigma\!=\!0.3$ for yaw and $0.15$  for pitch) are presented in Fig.~\ref{fig:more_results_3} and Fig.~\ref{fig:more_results_4}.

\subsection{More $360^\circ$ Results}

\begin{figure*}[t]
	\centering
	\includegraphics[width=1\textwidth]{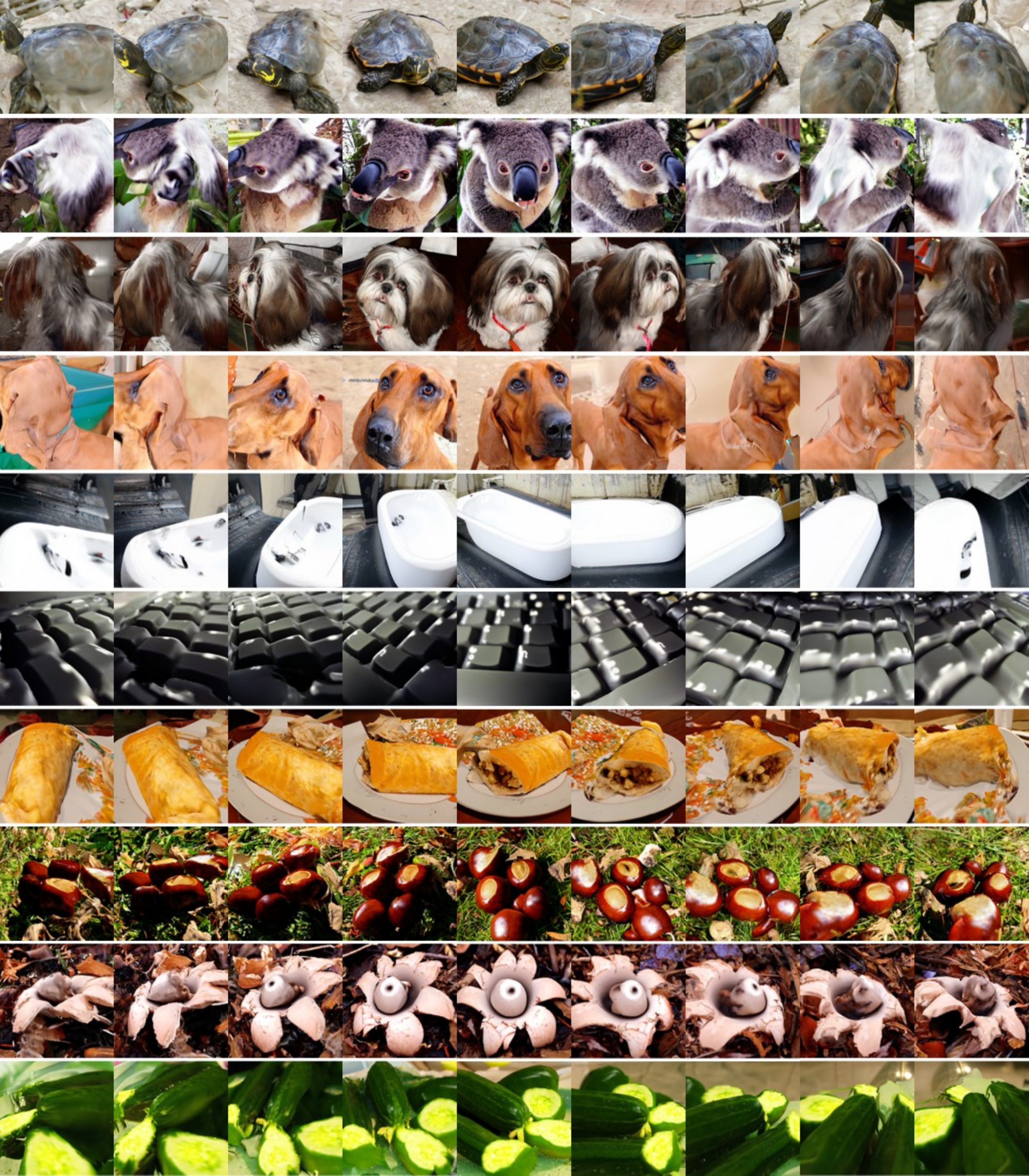}
    \vspace{-14pt}
	\caption{More $360^\circ$ generation results on ImageNet.}
	\label{fig:more_360_results}
    \vspace{-5pt}
\end{figure*}

In Fig.~\ref{fig:more_360_results}, we show more cases where our method successfully generates the results under a 360$^\circ$ camera trajectory.

\subsection{More $256^2$ Results}

\begin{figure*}[t]
	\centering
	\includegraphics[width=1\textwidth]{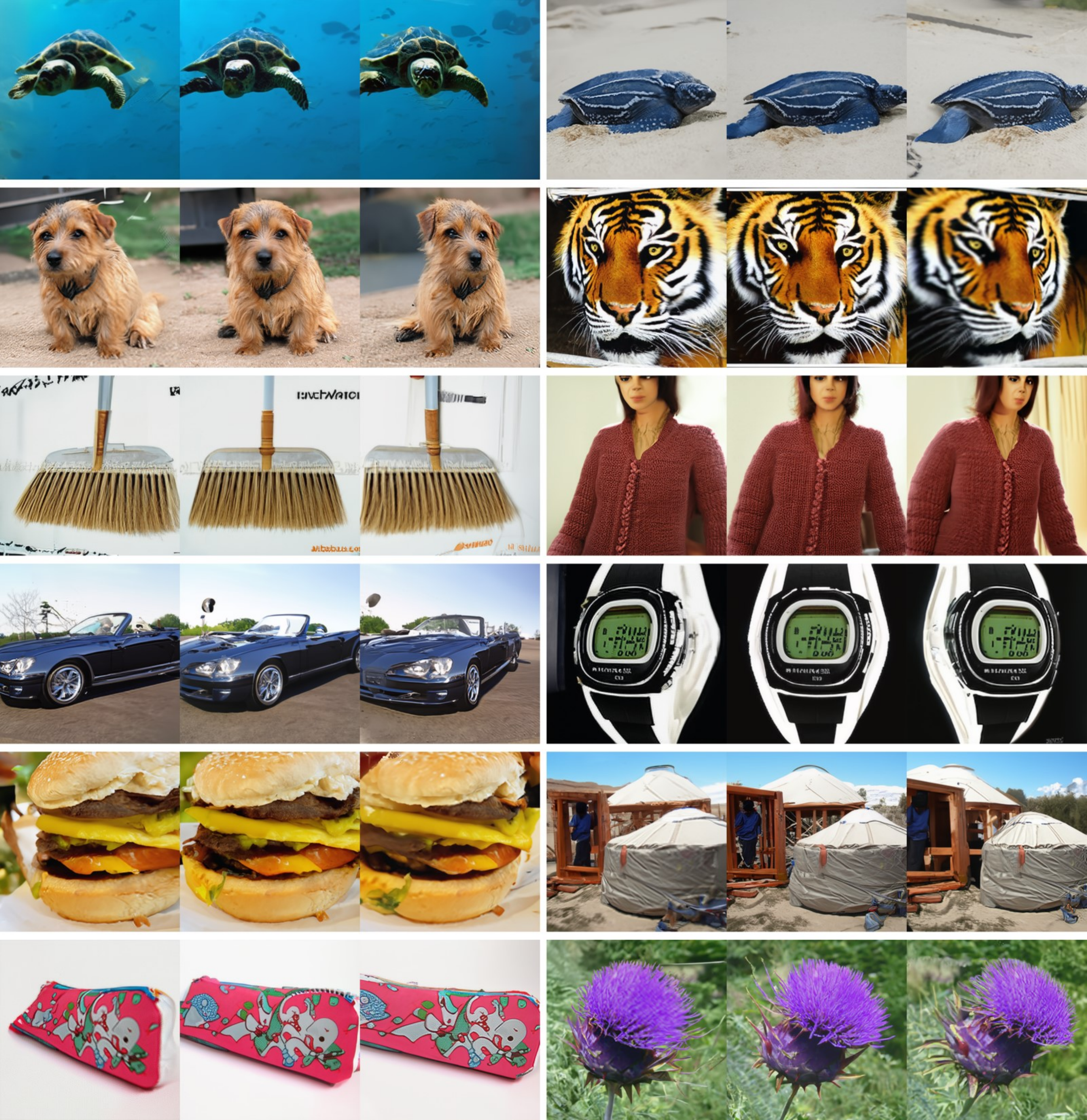}
    \vspace{-14pt}
	\caption{More $256^2$ generation results.}
	\label{fig:more_256_results}
    \vspace{-5pt}
\end{figure*}

In Fig.~\ref{fig:more_256_results}, we show more $256^2$ multiview results generated with the diffusion-based image upsampler.

\subsection{Shape Extraction}
\begin{figure*}[t]
	\centering
	\includegraphics[width=1\textwidth]{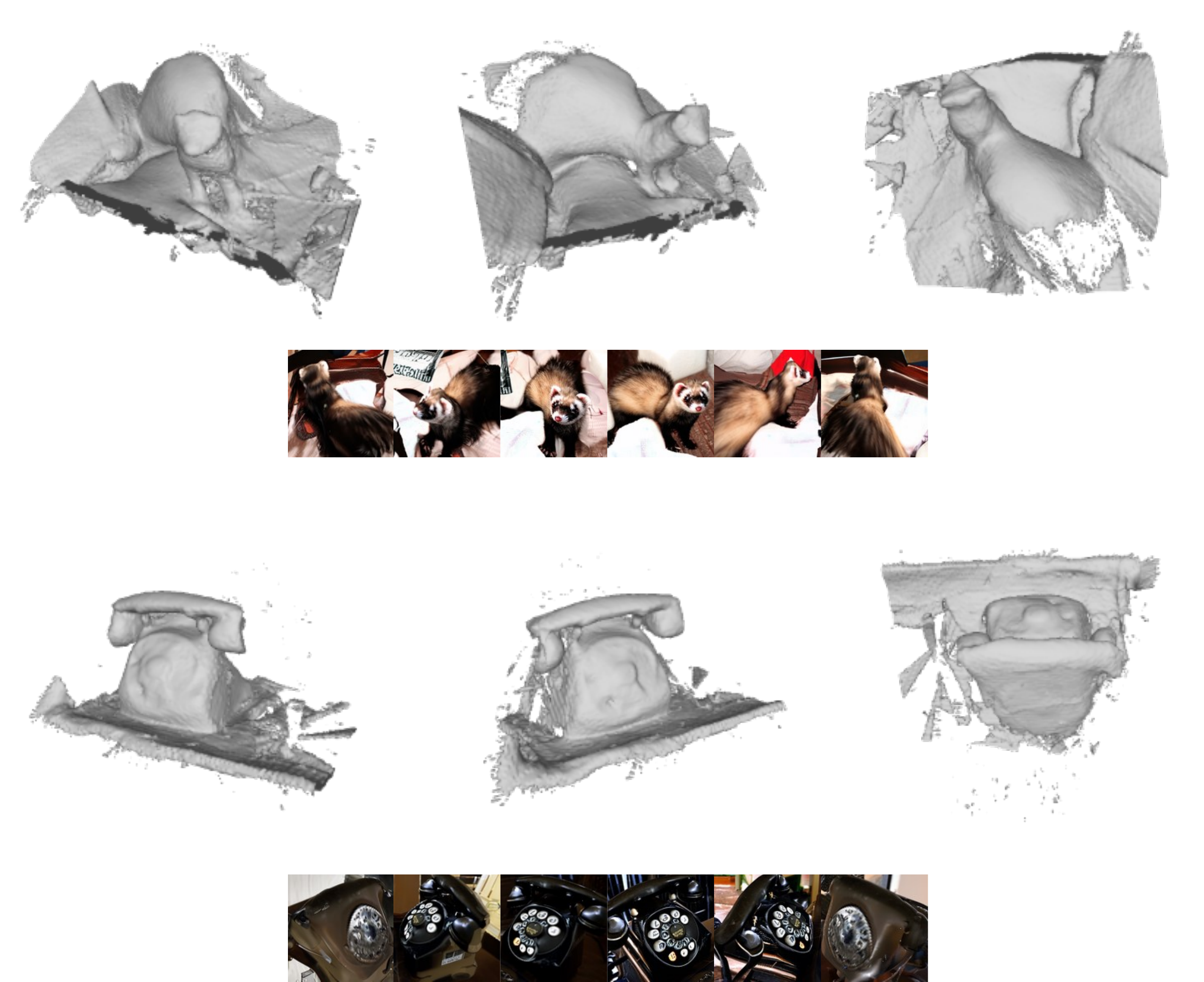}
    \vspace{-14pt}
	\caption{Shapes extracted using tsdf fusion and marching cubes.}
	\label{fig:tsdf_fusion}
    \vspace{-5pt}
\end{figure*}

To extact the 3D shape of a generated instance, we employ \emph{tsdf-fusion}~\cite{newcombe2011kinectfusion} to fuse the generated multiview depth maps into a voxel grid and  obtain the surface mesh using \emph{marching cubes}~\cite{lorensen1987marching}. Some examples are visualized in Fig.~\ref{fig:tsdf_fusion}.

\subsection{Failure Cases}

\begin{figure*}[t]
	\centering
	\includegraphics[width=0.7\textwidth]{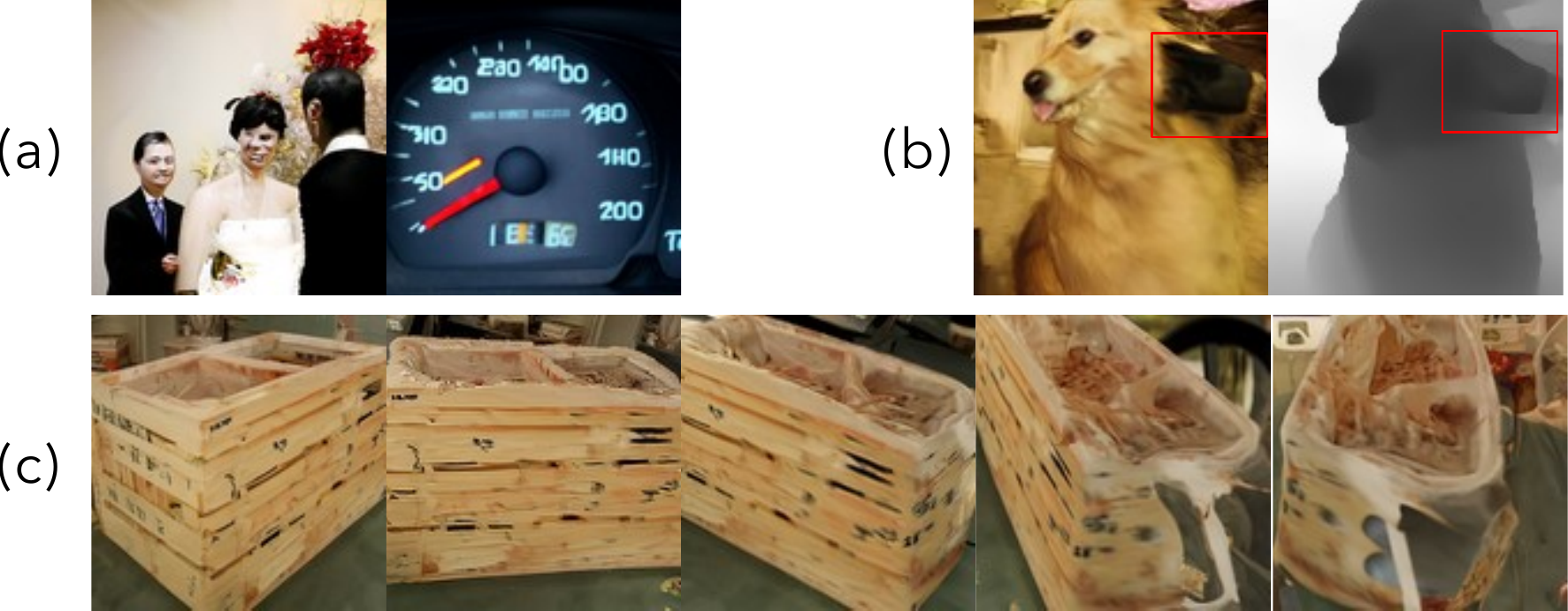}
    \vspace{-6pt}
	\caption{Failure cases. \textbf{(a)} Some  structures that are not well modeled by the network; \textbf{(b)} Severe mismatch between the generated color and depth map along occlusion boundary leads to poor novel view generation results.
    \textbf{(c)} Samples can be out of domain for very large views.}
	\label{fig:failure_cases}
    \vspace{-5pt}
\end{figure*}

In Fig.~\ref{fig:failure_cases}, we demonstrate some typical failure cases of our method.  First, our unconditional generation model $\mathcal{G}_u$ sometimes failed to model complex structures, which can be attributed to both limited model capacity and limited data for some object categories in ImageNet. Second, our model may generate severely-mismatched color and depth maps along occlusion boundaries which cannot be handled our texture erosion strategy. Under such situations, the conditional model will fail to generate proper contents under novel views.  For the $360^\circ$ generation, the sequence may not converge and the results can be completely out of domain.

\section{Limitations and Discussion}\label{supp:discussion}

Our current method has several limitations. First, the depth maps used for training are obtained by applying an existing monocular depth estimator~\cite{ranftl2020towards}. The depth error and bias in the data will inevitably affect the geometry quality of our generated results. How to alleviate such a negative impact or eliminate the requirement of depth (\eg, using multiview images) are left as our future work.

Our method also suffers from degraded image quality for large views, as mentioned in Sec.~5.3 of the main paper. There are at least two causes for this issue: domain drift and data bias. The errors on the generated depth will lead to distortions in the warped novel views, which will be accumulated and amplified in the iterative view sampling process and hence gradually drive the sample away from the real image distribution. Poor results will be generated when severe domain drift happens. Additionally, the object poses in the training datasets are usually not distributed uniformly. Side and rear views are often much less than frontal ones, rendering side- and rear-view image generation difficult.

Similar to most diffusion models, the image generation speed of our method is limited. However, we posit that these limitations can be gradually alleviated with the development of DM sampling accelaration~\cite{song2020denoising, salimansprogressive, ludpm}.

\end{document}